\documentclass[preprint,12pt]{elsarticle}
\usepackage{url}
\usepackage{color}
\usepackage{graphicx}
\usepackage{xspace}
\usepackage[bookmarks=false,hidelinks]{hyperref}
\usepackage{balance}
\usepackage[absolute,showboxes]{textpos}
\usepackage{amssymb}
\usepackage{lineno}
\usepackage{hyperref}
\usepackage[tikz]{bclogo}
\usepackage{subfig}
\usepackage{algorithm}
\usepackage{algorithmic}
\usepackage{amsmath}

\journal{Knowledge-Based Systems}

\begin{document}

\begin{frontmatter}
\title{Optimizing Hearthstone Agents using {an Evolutionary Algorithm}}
\author[uca]{P. Garc\'ia-S\'anchez}
\author[inra]{Alberto Tonda}
\author[uma]{Antonio J. Fern\'andez-Leiva}
\author[uma]{Carlos Cotta}
\address[uca]{Dept. of Computer Engineering, University of C\'adiz, Spain}
\address[inra]{UMR 782 GMPA, INRA, Universit\'{e} Paris-Saclay, Thiverval-Grignon, France}
\address[uma]{Dept. Lenguajes y Ciencias de la Computaci\'{o}n, Universidad de M\'alaga, Spain}

\begin{abstract}
Digital collectible card games are not only a growing part of the video game industry, but also an interesting research area for the field of computational intelligence. This game genre allows researchers to deal with hidden information, uncertainty and planning, among other aspects. This paper proposes the use of evolutionary algorithms (EAs) to develop agents who play a card game, Hearthstone, by optimizing  a data-driven decision-making mechanism that takes into account all the elements currently in play. Agents feature self-learning by means of a competitive coevolutionary training approach, whereby no external sparring element defined by the user is required for the optimization process. One of the agents developed through the proposed approach was runner-up (best 6$\%$) in an international Hearthstone Artificial Intelligence (AI) competition. Our proposal performed  remarkably well, even when it faced state-of-the-art techniques that attempted to take into account future game states, such as Monte-Carlo Tree search. This outcome shows how evolutionary computation could represent a considerable advantage in developing AIs for collectible card games {such as Hearthstone}. 
\end{abstract}


\end{frontmatter}

\section{Introduction}
\label{sec:intro}

Card games have been linked to artificial intelligence research since its inception. Classic games, such as Poker, have been highly studied due to their peculiarities, such as hidden information or discrete states. On the other hand, Collectible Card Games (CCGs) such as {\em Magic: The Gathering} offer a greater challenge, dealing not only with a wider search space, but {also} with their unique features: {each} card in these games features different rules/behaviors, so it can completely alter a game and even produce unpredictable combos, making collecting cards and designing decks one of its biggest strengths. Each created deck can have very different behaviors, making rich and complex gameplays emerge, {and hence} designing agents to play these games is not a trivial task.

One of the most famous Digital CCGs is {\em Hearthstone: Heroes of Warcraft}, which currently has over 40 million downloads \cite{PopularDCCGs}. In this game, two players, each using a deck designed before the game, employ combinations of cards (spells and minions) to remove life points from the opposing player, until one of {them} reaches 0 life points and is defeated. Due to the large number of cards available to create the decks, roughly 1,800, it is very complex not only to design these decks, but also to develop agents able to play various types of decks against a variety of enemy decks.

In this paper, we propose a method to automatically calculate the weights of a hand-coded agent that plays Hearthstone. A function to score all possible actions from a specific moment during the agent turn has been proposed, and optimized using an Evolutionary Algorithm (EA). As no other intelligent agents were available during its development, we used a {\em competitive coevolutionary} approach to assess the quality of the evolved agents. That is, during the evolution of one agent, the other individuals of the population (i.e., other agents) were used to calculate its fitness, by playing games against each other using different combinations of decks. This allows us to generate an agent versatile enough to confront a large number of behaviors.

The results show that our methodology can not only generate different types of agents, but also be able to win against other AI techniques. In fact, the best agent generated by our algorithm finished in second place (out of 33 contestants) in the first Hearthstone AI competition held in the Computational Intelligence in Games (CIG) conference 2018.

The main contributions of our work are the following: first, it proposes an evolutionary computation-based approach that can be used to optimize AIs to play CCGs. In particular, our proposal allows the optimization of a specialized automated system whose design is led by experienced players and whose performance depends on a number of parameters that are difficult to tune.  Moreover, one of our evolved agents has been proven to be very efficient in practice, performing remarkably in a competition featuring different types of AIs. Second, we propose to employ a competitive coevolutionary approach that can be used in a single population of an EA to lead the search {for better} solutions. This approach is especially interesting when there is uncertainty and hidden information in the problem and when there is no clear objective to optimize, but just the rather abstract concept of getting better at the game. Finally, our proposal can be seen as an alternative to the state-of-the-art method (i.e., the Monte Carlo Tree Search) currently employed to govern the behavior of virtual players in collectible card games.

The rest of the paper is structured as follows. First, a brief background on CCGs and EAs is presented in Section \ref{sec:background} {for the sake of completeness}. 
In that section we also 
discuss related work in the area of DCCGs, including Hearthstone, and the application of {EAs} to agent creation. The proposed approach is described in Section \ref{sec:proposed-approach}. The experimental setup is given in Section \ref{sec:experimental}. Finally, the discussion of the results is addressed in Section \ref{sec:results}, while  conclusions and future works are presented {in Section \ref{sec:conclusions}}.

\section{Background and related work}
\label{sec:background}
This section is devoted to describe some concepts that will be used along the paper, to ease its reading. In addition, this section also provides a discussion on related work.

\subsection{Digital Collectible Card Games: Hearthstone}

Collectible Card Games (CCGs) are a type of turn-based game where players prepare a deck of cards to be played before the game, and where the {\em deck-building} process is a very important part of the game experience. The goal of this kind of games is usually to beat the opponent by using the created deck. Every card has specific rules, that are applied when the card is played, affecting the game state. Some examples of CCGs are {\em Magic: The Gathering} \cite{Cowling12MagicMC}
 or {\em Pok\'emon Trading Card Game} \cite{Pokemon18}. Players must deal with hidden information in the form of the opponent player's hand, or the rest of his/her deck not played yet.

However, Digital Collectible Card Games (DCCGs), have some differences due to their nature: the possibility to add stochasticity to the play (via random effects encoded in cards), a more limited action set, and the modification of the rules by the developers in any time. Although there exist a number of well-known games, such as {\em Gwent}, or {\em The Elder Scrolls: Legends}, {\em Hearthstone: Heroes of Warcraft} is, without any doubt, the most famous and played DCCG nowadays \cite{PopularDCCGs}.

Hearthstone is a 2-player turn-based online DCCG launched in 2013 by Blizzard Entertainment. The game is based on the following elements:

{\em Card pool and decks}. Initially, the game provides a pool of cards that are available to the players (or {\em Heroes}). Then, at the beginning of a match, players are made to build a deck of 30 cards from the card pool. Note the difficulty of searching for an optimal deck due to the huge number of possibilities. A primary part of the success of Hearthstone is based on a policy of card expansion that, basically, means that about three times a year, 135 new cards (on average) are added to the card pool (this addition is called game expansion). Till September 2018, 13 expansions were produced. If one adds the cards that became part of the hall of fame of the game and the classic and basic sets, we obtain about 1,717 collectable cards. Moreover, at the time of playing, the Heroes (i.e., players) can have two copies of each card, except for the {so-called} legendary ones. This means that the number of cards that each player can consider, at the beginning of a match to construct his/her deck, is about 3,186. Therefore, in September 2018, the search space to construct the initial card deck for each player was $\binom{3,186}{30}$.
Two new expansions\footnote{\url{https://Hearthstone.gamepedia.com/}} were launched on December 2018 and  April 2019 so that currently Hearthstone features roughly 2,577 fully playable cards, of which exactly 2,005 cards can be collected by players (the others are generated by in-game effects). This gives an idea of the huge search space to optimize the card deck.

From this set of available cards, players are building their own collections by purchasing booster packs or receiving rewards as the player progresses through the game.

{\em Heros' health}. Each player (i.e., Hero) has 30 Health points at the beginning of the match. 

{\em Mana}. This is a resource that allows Heroes  to use (and apply) the cards and the Hero Powers. Basically, each card in the game has a mana cost that must be paid to use it. Each player has one Mana pool (i.e., an amount of Mana) and this has to be wisely managed by the player. Note that Mana is the only resource in the game and is the primary limiting factor on the play of cards. The supply of mana is represented by Mana Crystals.
At the starting of a match, the Mana pool of each player contains  1 Mana crystal. Then, in each turn this pool is increased with 1 more Crystal, up to a maximum of 10.

\begin{figure}[!t]
\centering
\includegraphics[scale=0.50]{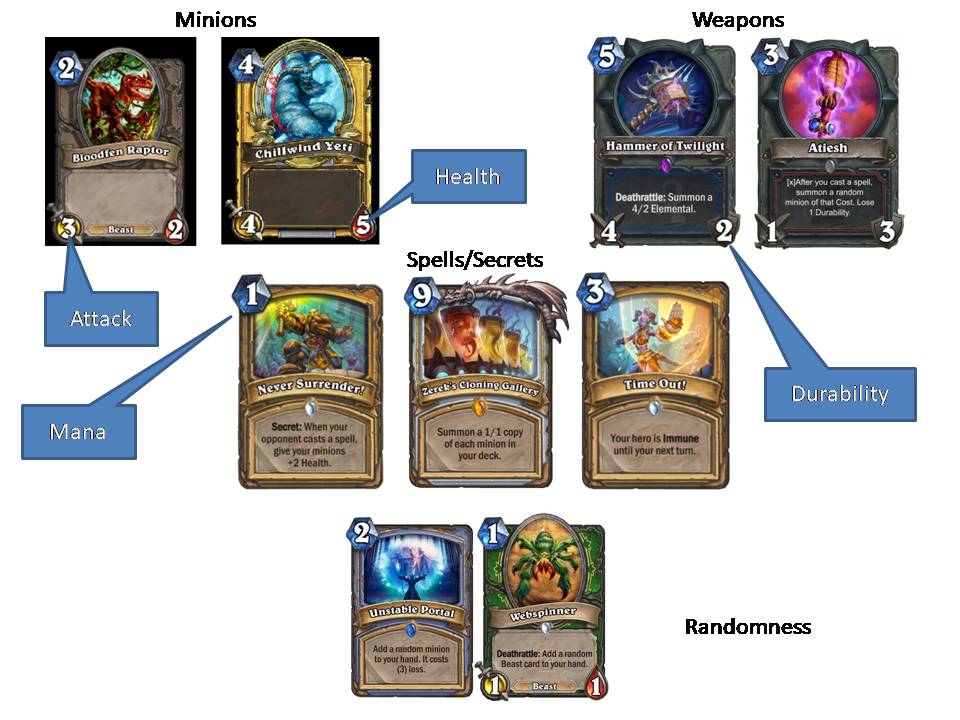}   
\caption{Examples of Hearthstone cards}
\label{fig: Hearthstone cards}
\end{figure}

{\em Type of cards}. There are several types of cards\footnote{Other card types -- namely {\em Hero cards} -- were introduced in the latest expansions, but they are out of the scope of the current work.}: {\em spells}, {\em minions} and {\em weapons}. In general, a card can have points associated to its  
Attack and Health attributes (shown in its bottom left corner and bottom right corner respectively), and a Mana cost (shown in its top left corner). See Figure \ref{fig: Hearthstone cards}. 

\begin{itemize}
\item {\em Spells} affect the battlefield by triggering a one{-}time effect or ability, and are discarded when used, with the exception of {\em Secrets}, that are placed next to the hero (not {on} the battlefield). The effect of a Spell is activated depending on some condition. Usually, Spell cards do not have Attack or Health attributes, only a mana cost. Spells provide functions ranging from simple damage-dealing and removal of minions, to providing useful enhancements, drawing cards, summoning minions and restoring Health.

\item {\em Minions} are persistent creatures placed on the battlefield that will help their Hero in the fight against the enemy. 
Minions have Health points (MH) and Attack points (MA), but can also possess special abilities, such as {\em Charge} (the minion can attack immediately after being placed on the battlefield) or {\em Inspire} (activate an effect when the Hero Power is used). Minions are controlled by the player who summoned them, and can be commanded to attack their opponent's minions, or even the opposing hero. Certain minions, with the \emph{Taunt} ability, can act as defenders, preventing enemies from attacking other friendly minions or the controlling Hero until these minions are removed.
Minions are a major element in battles between Heroes, and are usually responsible for the majority of all damage dealt in a game.

\item Finally, {\em Weapons} are special cards which can be equipped by Heroes and allow them to attack other characters. Each weapon has an Attack value and a Durability value (number of times it can be used to attack). Weapons are the main way of allowing heroes to attack other characters.
\end{itemize}

In general, when a minion's health (resp. weapon's Durability) is reduced to zero, the minion (resp. weapon) is destroyed.

Note that many cards in the game deal with random actions. As a consequence, randomness (and imprecise information) is an important feature of this game that has to be managed by experienced players{, thus making} the game even more interesting from an optimization point of view.

Regarding the mechanics, this is a 2-player turn-based game that can be seen as a sequence of turns. The game mechanics 
are as follows: initially, each player chooses 30 cards from his/her collection (or uses a previously saved deck). The goal of the game is to reduce the enemy Hero's health points from 30 to 0 by making use of the cards in the Player's card deck. At the beginning of each {player's} turn, the Mana Pool is refilled, and a new crystal is obtained ({the} initial size {of the pool} is 1, and {its} maximum size is 10). Possible actions in a turn are: play a card in hand, use the Hero Power, attack with a minion on the battlefield, attack with a weapon, or end turn. During each turn, the player can spend crystals from his/her Mana pool to play a {\em Spell Card}, to execute the {\em Hero Power} (once per turn), to equip a {\em Weapon}, or to put a {\em Minion Card} {on} the battlefield. Minion cards already on the battlefield can be used to attack other minions or the enemy Hero. Attacking with a minion does not consume crystals. Minions have health points and attack points, and they can (normally) attack once per turn. If a minion attacks another minion, the health of the other is reduced depending on the attack points, and vice versa. Minions are killed when they reach 0 health points. Minions can have different abilities that affect their {behavior}: {\em Stealth}, {\em Taunt}, {\em Windfury}, etc. If the Hero has an equipped Weapon, it can also attack other minions or the enemy Hero, once per turn. The number of turns a Weapon can be used before it breaks depends on {its} durability, and only one Weapon can be equipped at the same time. Heroes can use the {\em Hero Power} of their Hero Class once per turn. For example, the {\em Hunter} can spend 2 Mana crystals to inflict 2 direct damage to the enemy Hero. Some actions can activate the {\em Secrets} placed next to the Hero. For example, the Secret {\em Vaporize} destroys the first minion that attacks the Hero. Players can also end their turn at any point, even if they still have possible actions to perform. For example, they can save a card for later, even if they have enough Mana to play it in the current turn. The game continues, turn by turn, until one of the two players (Heroes) kills the other by reducing his/her health points to 0.

As already mentioned, because of the wide variety of cards, players can create decks with a large number of different behaviors. Normally, these behaviors can be reduced to the following archetypes: Aggro, Combo or Control. Using an {\em Aggro} (aggression) deck the player tries to finish the game as soon as possible using low-cost cards. On the other hand, the objective of {\em Combo} decks is to survive until they have an optimal combination of cards that allows them to release a large amount of damage, usually during a single turn. Finally, {\em  Control} decks are based on eliminating the first threats of the game until later turns and then playing more powerful cards (at a higher mana cost). We refer the reader to \cite{Garcia18Knowsys} for further additional information on Hearthstone.

\subsection{Evolutionary Computation}
\label{subsect: EA and CEA}

Evolutionary Computation (EC) is a scientific field that includes a large number of bio-inspired techniques. Belonging to this field, Evolutionary Algorithms (EAs) are stochastic optimization methodologies \cite{DBLP:series/ncs/EibenS15}. 

EAs are inspired by the {\em natural selection} process, using the concept of {\em fitness} to score each solution (also called {\em individual}). Fittest individuals have a higher probability to reproduce and generate new solutions that will inherit part of their structure. After a certain number of iterations, where individuals will recombine between them to form new individuals, it is expected that the selective pressure will produce better solutions.

At each iteration, or {\em generation}, different operators are applied to parent individuals to recombine them and generate new offspring ({\em crossover}), or to modify existent ones ({\em mutation}). At the end of each generation, the least fit individuals are removed. The process continues until a termination criterion is met \cite{FernandezStop15}.

One of the advantages of EAs is that they can obtain optimal, or near-optimal solutions, {in problems for which these are} hard to find for human experts and, on the other hand, they are able to deliver such solutions in a reasonable amount of time, even when facing problems with high dimensionality, where traditional optimization techniques usually fail. 

Among the different types of EAs, Genetic Algorithms (GAs) are the most {well-known} \cite{DBLP:series/ncs/EibenS15}. However, depending on the encoding of the candidate solutions, other {flavors}, such as Evolutionary Strategies (ES) \cite{rechenberg1978evolutionsstrategien,schwefel1977numerische} can be used. ES are better suited to deal with solutions codified as a vector ({\em genome}) of floating-point values. The key difference with respect to other EAs is that each solution also encodes a $\sigma$ value for each element on the vector, that defines how the mutation on this specific gene will be applied. ES also deal with {the} replacement of the individuals in different ways, for example, producing $\lambda$ {offspring} from a population of $\mu$ {individuals} and keeping the best $\mu$ individuals {after} combining both groups $(\mu+\lambda)$.

The fitness of an individual is computed by applying a {certain objective} function 
to the values of the genome, the candidate solution{,} to measure how far or near the individual is from the optimal solution of the problem. In the case of game AI optimization, the fitness of an individual is usually calculated by letting the candidate (indistinctly termed in this paper as agent, bot, or game AI) play in a game simulator, and obtaining some metrics of the game as the fitness value (for example, the score of the match, or the number of victories \cite{MoraFGGF12}). This can be useful if we want to improve a game AI against other available AIs. 

Coevolution is another interesting evolutionary approach that has been used with success in videogame development.
The concept of coevolution involves individuals which interact with each other and that might belong (or not) to the same species (basically populations). From a general point of view, the classical model of coevolution proposes to coevolve populations belonging to the same species. This means that all the individuals have the same genetic structure or codification, and from this idea, two approaches can be identified. The first approach uses a unique population and the evaluation process is carried out by having individuals face each other according to a selection mechanism. A direct consequence is that here reproductive relationships emerge as a natural process. The second case employs different populations, 
and tries to mimic (and exploit the features of) an arms race between coevolving populations that belong to the same species (or at least to the same biotic niche), namely strategies, rules, tracks for racing, or any other. Note that, in both approaches, a competition is usually present, so that the model can also be named \emph{competitive coevolution} \cite{dawkins1979arms}. In literature, it is possible to find a number of coevolutionary approaches applied to games, such as Tic Tac Toe \cite{DBLP:conf/icga/AngelineP93},  pursuit-evasion games  \cite{reynolds1994competition}, predator-prey games, \cite{DBLP:journals/alife/Sims94},  Real-Time Strategy games \cite{Fernandez-AresG16}, or capture-the-flag games 
\cite{DBLP:conf/cig/SmithAHL10}, just to mention a few. 

Other forms of coevolution, not considering just one single species, are also possible. 
In fact, some coevolutionary models are based on multispecies interaction. Sometimes these species evolve by competition, but they can also coexist cooperatively, that is to say, in a mutually beneficial relationship \cite{MaAndOthers:CooperativeCoevolution2019}. However, this type of coevolution based on multispecies is less common in games \cite{DBLP:conf/cec/CardonaTNB13,nogueira14virtual}.

\subsection{Optimization of digital collectible card games}
\label{sec:sota}

Since the nineties, card games have been an important research field, starting with digital versions of classic games, such as Poker \cite{Broeck_Poker_MCTS_2009} or Solitaire \cite{Bjarnason_Solitaire_2009}. Recently, with the increased interest in CCGs, a new testbed for AI research has appeared, implying new aspects to study, such as hidden information and uncertainty \cite{FRANK_incomplete_info_cards_1998}. Moreover, given the huge amount of cards present in this kind of games, a large number of effects can appear --some of them even unforeseeable-- that can affect the current state of the play.

Recently, CCGs have gained more presence in the computational intelligence research community, due to the appearance of new simulation software {that is} able to test different decks and autonomous agents. Some examples are  Apprentice\footnote{\url{http://apprentice.nu/}} or Magic Workstation\footnote{\url{http://www.magicworkstation.com/}}, which let the player manage card collections and play against other (human) players online. Even more recently, Hearthstone simulators such as \textit{MetaStone} \cite{METASTONE_2017} and {\em SabberStone} \cite{SABBERSTONE} have arisen as a real option for testing AI approaches. 

One of the first works on the subject, by Mahlmann et al. \cite{MahlmannDominion12}, is a complete study on balancing the {\em Dominion} card game using EAs. In this game, decks are formed by stacks of copies of a set of cards, placed at the beginning of the game, with different sets in each session, and therefore, forcing players to adapt their strategies. Each individual of the EA is a vector of 10 cards (from a pool of 25 available). A similar approach was proposed by Garc\'{\i}a-S\'anchez et al.  \cite{Garcia18Knowsys,Garcia16CIG} with application to Hearthstone, but using a vector of 30 cards (deck size in the game) from a pool of more than 700. 
%%%
The objective here is to search the optimal combination of cards (i.e., the optimal deck) that optimizes the performance of a specific (and fixed) game AI (in this case the default AI that comes with the simulation software). Therefore, the candidates or individuals to be optimized represent decks. Nine decks, one per each hero class, were optimized using an EA with a smart mutation operation, while the Metastone software simulated the matches of the game. When played by the agent, the so-obtained decks outperformed the best hand-made decks created by humans. As in this paper, the number of victories against a wide set of enemy decks was used to calculate the fitness value of a candidate deck. A similar approach was presented by Bhatt et al. \cite{Bhat18Exploring}, but feeding the difference on health of the players at the end of the game (rather than the number of victories) to a sigmoid function for fitness purposes. However, in the two previous works the AI used in the simulations was limited to the default agent.

Hearthstone has also been used as a case study for other aspects unrelated to AI. For example, 
Wanderley et al. \cite{WanderleyHoningstone16} presented a method to generate unexpected and original combinations of cards, following a creativity-focused method. This work only tries to generate `uncommon' card combinations, calculating efficiency and rarity metrics from a previously built database of combos, extracted from human gameplays. 

Note that
all previous works were focused on the deck-building aspect of DCCGs. Currently, another line of research is pushed. This line basically consists of designing and implementing automated proposals (usually based on {advanced} AI methods) to control the decision-making mechanism of a virtual player. For example Monte Carlo Tree Search (MCTS) has been applied to deal with the imperfect information of the game \emph{Magic: The Gathering} \cite{Cowling12MagicMC}. The authors assumed that the hidden and random information is known by the players (a procedure called \emph{determinization}) to develop an advanced MCTS approach. This method was not only {able} to win against a human-expert heuristic system, but {it was also claimed that it could even} outperform strong human players. Information Set MCTS has also been used to play the Pok\'emon card game \cite{Pokemon18}, obtaining better results than the standard MCTS method.

As Hearthstone was created to be played digitally (unlike \emph{Magic: The Gathering}), it started to be a {\em de-facto} testbed for this kind of games.
In fact, the game not only presents hidden information and a wide branching space as in Magic, but also offers clear and defined actions, not to mention stochastic outcomes (particularly evident by the random factors of the game). In one of the first papers that addressed the generation of virtual players to play this game, Bursztein  \cite{Bursztein_Legend_Hearthstone16} described a statistics-based agent. This agent applies learning methods to predict, with a high accuracy level, the cards that its opponent will play in the following turns. This accuracy decreases with time, as expected, as the number of options available increases.

Nowadays, MCTS has become the state-of-the-art method to implement the  mechanism that leads the decisions of virtual players in   card-games \cite{Swiechowski18}. 
The first works applying it to Hearthstone
proposed the use of MCTS \cite{Santos17}, or its combination with neural networks \cite{Zhang17}. Recently, Swiechowski et al. \cite{Swiechowski18} modified MCTS with different methods to handle randomness and imperfect information, with machine learning from datasets to create the scoring functions. However, none of the previous works compared the generated bots against complex ones created by other researchers, focusing only on random-movement or greedy-based agents to measure the performance. In this sense, the comparison of different game AIs (i.e., the decision-making mechanisms of the virtual players) to play a DCCG is really important in order to assess their adequacy and, mainly, their efficiency. Moreover, when an evolutionary algorithm is employed to generate the game AI, this issue (i.e., to measure the `goodness' of the candidate solutions) is crucial, as the search is guided by the efficiency of the candidate to play  the game, and this can only be measured in a match against other (perhaps virtual) players. %
This is a problem that has to be addressed when there are no other methods with which to compare or {it} is difficult to implement them. Next section describes our proposal to cope with this issue in the problem of optimizing
hand-coded strategies to efficiently play HeartStone.

\section{Generating Virtual Players for Hearthstone}
\label{sec:proposed-approach}

This section describes our proposal to automate the generation of efficient decision-making mechanisms to play Hearthstone, that is to say, our method to generate efficient game AIs.  

Our approach consists of a coevolutionary algorithm that  bootstraps the performance of agents playing Hearthstone. The core functioning of these agents will be described in more detail in next subsections. Firstly, in Section \ref{subsect: agent description}, we describe a data-driven specialized automated system, that was constructed manually, to play the card game. The performance of this system depends not only on the knowledge provided to the algorithm designer by an experienced human player, but also on the specific values assigned to a high number of parameters. Then, in Section \ref{subsec:proposed-evolutionary-algorithm} we present our coevolutionary proposal and dive into the concrete details of the difficulty to evaluate the candidates and our suggestion to deal with the fitness evaluation.

\subsection{Agent Description}
\label{subsect: agent description}

Our proposed agent ($A$) {is} implemented using the {\em SabberStone} framework {(cf. Section \ref{subsec:metastone}) and} mimics a virtual player that evaluates how the execution of any possible action 
(i.e., play card in hand, attack with a minion/weapon or end turn)
that can be executed {at a given} moment affects the state of the game. The objective after carrying out this evaluation phase is {having} the virtual player execute the action that provides the best performance (in terms of changing the current state to the best possible scenario in the game). These two steps are repeated until the action `end turn' is executed. 
{One important issue is that our agent is a data-driven mechanism in which the election of the best action to take strongly depends on the values initially given to a set of 21 weights in the form of $\vec{w}={\langle \mathbf w_1},\ldots,{\mathbf w_{21}} \rangle$. Changing the values in $\vec{w}$ will affect the performance of the agent, as it is explained below.}

\begin{table}[!t]
\caption{Weights used to compute the score of an action.}
\label{tab:weightsBattlefield}
\resizebox{\textwidth}{!}{
\begin{tabular}{cclp{5.8cm}}
\hline
\textbf{~Id~} & \textbf{~Acronym~} & \textbf{Name} & \textbf{Description} \\ \hline
\\
\multicolumn{4}{c}{\textsf {Weights to score the difference in Heroes stats $\Delta^{a,S}_{attributes(hero)}$. See Eq. (\ref{eq:health})}} \\ \hline 
{$\mathbf w_1$} & HHR & Hero Health Reduced &  Difference in health and armor after executing the action \\ 
{$\mathbf w_2$} & HAR & Hero Attack Reduced &  Difference in attack after executing the action \\ 
\hline 
\\
\multicolumn{4}{c}{\textsf {Weights to score $\Delta^{a,S}_{minions(hero)}$ after a change {on} the battlefield.  See Eq. (\ref{eq:state of minions})}} \\ \hline 
{$\mathbf w_3$} & BMHR & Minion Health Reduced &  Difference in health \\ 
{$\mathbf w_4$} & BMAR & Minion Attack Reduced &  Difference in attack \\ 
{$\mathbf w_5$} & BMA  & Minion Appeared & A new minion appeared {on} the battlefield\\  
{$\mathbf w_6$} & BMK  & Minion Killed & A minion was killed\\  
\hline
\\
\multicolumn{4}{c}{\textsf {Weights to score $\Delta^{a,S}_{secrets(hero)}$. See Eq. (\ref{eq:secretmana})}} \\ \hline 
{$\mathbf w_7$} & BSR  & Secret Removed &  A secret has been removed/appeared \\ 
\hline 
\\
\multicolumn{4}{c}{\textsf {Weight to score $\Delta^{a,S}_{manaConsumed}$ See Eq. (\ref{eq:secretmana})}} \\ \hline 
{$\mathbf w_8$} & BMR  & Mana Reduced & Mana reduced after executing the action\\ 
\hline
\end{tabular}
}
\end{table}

\begin{table}[!htb]
\caption{Weights used to calculate the value of a Minion via the function $value\mathit{Of}(m)$.
}
\label{tab:weightsMinion}
\resizebox{\textwidth}{!}{
\begin{tabular}{cclp{5.6cm}}
\hline
\textbf{~Id~} & \textbf{~Acronym~} & \textbf{Name} & \textbf{Description} \\ \hline
{$\mathbf w_9$} & MH   & Minion Health & Current health of the minion\\  
{$\mathbf w_{10}$} & MA   & Minion Attack & Current attack of the minion\\  
{$\mathbf w_{11}$} & MHC  & Minion Has Charge & Minion can attack the turn it enters into play\\ 
{$\mathbf w_{12}$} & MHD  & Minion Has Deathrattle & Minion does something when dying\\  
{$\mathbf w_{13}$} & MHDS & Minion Has Divine Shield & First attack do not harm the minion\\  
{$\mathbf w_{14}$} & MHI  & Minion Has Inspire & Does something every time the hero power is performed\\  
{$\mathbf w_{15}$} & MHLS & Minion Has Life Steal & The health removed to an enemy by this minion is gained by the hero\\  
{$\mathbf w_{16}$} & MHS  & Minion Has Stealth & Cannot be target of spells and attacks until it attacks the first time\\  
{$\mathbf w_{17}$} & MHT  & Minion Has Taunt & The other minions cannot be attacked until the taunt is removed (or the minion is killed)\\  
{$\mathbf w_{18}$} & MHW  & Minion Has Windfury & Can attack twice\\  
{$\mathbf w_{19}$} & MHP   & Minion Has Poison & Kills after the first attack\\ 
{$\mathbf w_{20}$} & MR   & Minion Rarity & Rarity of the card: Common (1), Rare (2), Epic (3), Legendary (4) \\  
{$\mathbf w_{21}$} & MM   & Minion Mana Cost & Cost necessary to invoke the minion \\ \hline 
\end{tabular}
}
\end{table}

More formally, let 
{$\vec{w}={\langle \mathbf w_1},\ldots,{\mathbf w_{21}} \rangle$ be the initial set of weights initially preset in our agent, and } let $S$ be the current state of the game associated to a turn in which the virtual player has to make game decisions. This state $S$ is characterized by the 
information that can be gathered from the current scenario of the game. This information consists {of} a number of game data (e.g., amount of health and armor of the player, cards that are placed {on} the battlefield, amount of mana in the Mana Pool, cards that can be used, etc). Note that this information is also visible to human players.
Our agent first identifies the set $A=\{a_1,\ldots,a_n\}$ of all the $n$ possible actions that can be executed according to the current state $S$  of the game. Second, it compares how the execution of any action $a_i$ affects state $S$. This is done by measuring the differences {in} the game data (for instance, health and armor of both the player and the opponent, amount of minions {on} the battlefield, and/or mana consumed, just to mention some of them.) {as a result of} applying action $a_i$ in {that particular} state $S$.  
{Note that that the value of the weights in $\vec{w}$ have influence in the result of applying any action in state $S$.} Finally, the action $a \in A$ with the highest difference is selected to be executed. This action is considered as the action that affects more positively to the objectives of the player. 
This process is performed until no more actions in the turn are possible (e.g., the agent has no cards on the board and there is no enough mana in the pool to use other cards owned by the agent) or the `skip turn' action is selected because it obtained the higher score. 

In order to increase the comprehension of our proposal and ease  the coding (and replication) of our virtual agent, in the following we formally describe how to evaluate the difference {between} the state $S$ and the state that results after applying an action $a$ in the state $S$, 
{considering that $\vec{w}={\langle \mathbf w_1},\ldots,{\mathbf w_{21}} \rangle$ is the predefined set of weights in the agent}. This is done via the function $\Delta^{a,S,{\vec{w}}}$, defined as follows:

 \begin{equation}
\Delta^{a,S{,\vec{w}}} = \Delta^{a,S{,\vec{w}}}_{state\mathit{Of}(enemy)} - \Delta^{a,S{,\vec{w}}}_{state\mathit{Of}(agent)}- \Delta^{a,S{,\vec{w}}}_{manaConsumed} \\
\label{eq:score}
\end{equation}
{where $\Delta^{a,S{,\vec{w}}}_x$ denotes the difference between the numerical value of the parameter or function $x$ in state $S$ and its corresponding value after applying action $a$. For simplicity, we will frequently omit the superscript $\vec{w}$ and simply use $\Delta^{a,S}_x$ when the weights are implicit in the context.}

{Then}, for $hero \in \{enemy,agent\}$:

\begin{equation}
\Delta^{a,S}_{state\mathit{Of}(hero)} = \Delta^{a,S}_{attributes(hero)} + \Delta^{a,S}_{minions(hero)}+ \Delta^{a,S}_{secrets(hero)} \\
\label{eq:state of hero}
\end{equation}

Therefore, {considering that the agent have preset the values of the weights $\vec{w}={\langle \mathbf w_1},\ldots,{\mathbf w_{21}} \rangle$,}
$\Delta^{a,S{,\vec{w}}}${, as defined in Equation~(\ref{eq:score}),} basically evaluates how  executing action $a$ affects the states (before and after applying action $a$) of both the enemy and the virtual agent as well as to the amount of mana. The idea is to select the action that best serves to the objectives of the agent (i.e., the virtual player).  Note that the difference between the game states of a hero (i.e., enemy or agent) before and after applying action $a$ depends on variations associated with the values of its main attributes (e.g., health or armor), its minions placed {on} the battlefield and its secrets. This is reflected in Equation~(\ref{eq:state of hero}).

Equations~(\ref{eq:score}) and~(\ref{eq:state of hero}) were defined by an experienced player trying to cover all the aspects of the game, and depend on 21  parameters (shown in Tables \ref{tab:weightsBattlefield} and \ref{tab:weightsMinion}, and ranging in the real interval $[0.0,1.0]$) whose values can strongly influence the decision to select the best action to execute. 
 In the following we describe how the {different aforementioned} variations are calculated and their dependency on the different parameters.

{Firstly}, 
the variations of the attributes of the hero are calculated on three main values, namely, health, armor and attack damage, whose influence is determined by parameters {${\mathbf w_1}$ and ${\mathbf w_2}$:}

\begin{equation}
\begin{aligned}
\Delta^{a,S}_{attributes(hero)} = &\ \ \ {\mathbf w_1}\cdot ( \Delta^{a,S}_{health\mathit{Of}(hero)} + \Delta^{a,S}_{armor\mathit{Of}(hero)} ) \\
 & + {\mathbf w_2}  \cdot \Delta^{a,S}_{attackDamage\mathit{Of}(hero)}
\end{aligned}
\label{eq:health}
\end{equation}

The variations on health and armors are summed, whereas the amount of attach damage is considered separately.
Note that a reduction (i.e., variation) in the enemy health, armor  and attack increases the score of Equation (\ref{eq:score}) whereas a decrease in the same attributes of the agent decreases it. For example, executing an action $a$ of `Attack with a Minion with 2 Attack Points to Enemy Hero' will imply a change in Enemy Health from 20 (before executing the action) to 18 (after executing $t$), and this affects the value of $\Delta^{a,S}$.

Equation~(\ref{eq:state of minions}) measures the changes {on} the battlefield taking into account the modifications to the set of minions of each player. The differences in minions attack and health, the minions appeared or killed, and the specific value of every minion modified are used as parameters of the equation. This may imply giving more value to an action that decreases a powerful minion health, than killing a weak one, depending on the weights ${\mathbf w_3},\ldots,{\mathbf w_6},{\mathbf w_9},\ldots, {\mathbf w_{21}}${:}

\begin{equation}
\begin{array}{rll}
\Delta^{a,S}_{minions(hero)} = &  & {\mathbf w_3 } \cdot \Delta^{a,S}_{minionsHealth\mathit{Of}(hero)}\ \\
                               & +& {\mathbf w_4 } \cdot \Delta^{a,S}_{minionsAttack\mathit{Of}(hero)}\\
                               & +& {\mathbf w_5 } \cdot \Delta^{a,S}_{minionsKilled\mathit{Of}(hero)}\ \\
                               & -& {\mathbf w_6 } \cdot \Delta^{a,S}_{minionsAppeared\mathit{Of}(hero)}
\end{array}
\label{eq:state of minions}
\end{equation}

Note {that} the variations of the set of minions of each player after executing an action {depend} on the variations  in health and attack of the minions that are still {\em alive}, the value of those minions that have been annihilated/killed, and the value of the new minions that are positioned {on} the battlefield. This is formally defined below.

\begin{equation}
\begin{aligned}
\Delta^{a,S}_{minionsHealth\mathit{Of}(hero)} &= \sum_{m\ {\rm is}\ alive}\left[ \Delta^{a,S}_{health\mathit{Of}(m)} \cdot value\mathit{Of}(m) \right] \\
\Delta^{a,S}_{minionsAttack\mathit{Of}(hero)} &= \sum_{m\ {\rm is}\ alive}\left[\Delta^{a,S}_{attack\mathit{Of}(m)} \cdot value\mathit{Of}(m)\right]\\
\Delta^{a,S}_{minionsKilled\mathit{Of}(hero)} &= \sum_{m\ {\rm is}\ new}\left[value\mathit{Of}(m)\right]\\
\Delta^{a,S}_{minionsAppeared\mathit{Of}(hero)} &= \sum_{m\ {\rm is}\ killed}\left[value\mathit{Of}(m)\right]
\end{aligned}
\end{equation}

Every minion $m$ has an associated value ($value\mathit{Of}(m)$) calculated as the following scalar product :

\begin{equation}
\begin{aligned}
value\mathit{Of}(m) & = \langle {\mathbf w_9},\ldots, {\mathbf w_{21}} \rangle \cdot \vec{v}_{attributes\mathit{Of}(m)} \end{aligned}
\label{eq:minion}
\end{equation}

\noindent where $\vec{v}_{attributes\mathit{Of}(m)}$ is a vector containing the following 13 values:

\begin{equation}
\begin{aligned}
&\vec{v}_{attributes\mathit{Of}(m)} = \langle\  \\ 
& \ \ \  \mathit{healthOf}(m),\ \mathit{attackDamageOf}(m),\ \mathit{hasCharge}(m), \\
&\ \ \  \mathit{hasDeathrattle}(m),\ \mathit{hasDivineShield}(m),\ \mathit{hasInspire}(m), \\ 
&\ \ \  \mathit{hasLifeSteal}(m),\ \mathit{hasStealth}(m),\ \mathit{hasTaunt}(m), \\
&\ \ \  \mathit{hasWindFury}(m),\ \mathit{hasPoison}(m),\ \mathit{rarityOf}(m), \mathit{manaCost} \\
&\ \ \ \rangle
\end{aligned}
\label{eq:minion attributes vector}
\end{equation}

The functions with the pattern $HasAbility(m)$ return 1 if the minion $m$ has a specific special ability (or trait), and 0 otherwise. For example, $hasCharge(m)$ returns 1 if the minion $m$ has the  \emph{Charge} ability, and 0 otherwise. Note that this vector contains information related to the minion such as health, attack, rarity, and special abilities, to mention some of them so that $\vec{v}_{attributes\mathit{Of}(m)}$ group all the relevant information related to minion $m$. The influence of each of these 13 parameters depends directly in  the value of the 13 weights ${\mathbf w_9},\ldots, {\mathbf w_{21}}$ shown in Table~\ref{tab:weightsMinion}.

Finally, the score of the secrets appeared/removed and the mana used by executing the action $a$ in state $S$ are calculated as shown in Equation \ref{eq:secretmana}:

\begin{equation}
\begin{aligned}
\Delta^{a,S}_{secrets(hero)} &= {\mathbf w_7} \cdot \Delta^{a,S}_{secrets\mathit{Of}(hero)} \\
\Delta^{a,S}_{manaConsumed} &= {\mathbf w_8} \cdot \Delta^{a,S}_{mana\mathit{Of}(hero)} \\
\end{aligned}
\label{eq:secretmana}
\end{equation}

\begin{algorithm}[!t]
\begin{algorithmic}
\STATE \COMMENT{{This is the internal logic used by an agent, instantiated with a vector of preset weights $\vec{w}={\langle \mathbf w_1},\ldots, {\mathbf w_{21}}\rangle$, to obtain the best action to execute in a state $S$ of the game every time during the agent's turn. The turn finishes when the `skip turn' action is returned.} }
\STATE A $\gets \{a_1,\ldots,a_n\}$ (i.e., the set of all possible actions to execute in $S$)
\STATE bestAction $\gets$ `skip turn'
\STATE bestScore $\gets$ -$\infty$
\FOR{$i \in \{1,\ldots,n\}$}
	\IF{$\Delta^{a_i,S{,\vec{w}}} >$ bestScore}
	    \STATE	bestAction $\gets a_i$
	    \STATE bestScore $\gets \Delta^{a_i,S{,\vec{w}}}$
	\ENDIF
\ENDFOR
\RETURN bestAction
\end{algorithmic}
\caption{Select the best action.}
\label{alg:agent}
\end{algorithm}

The process to obtain the best action that our agent can {execute} in any state $S$  of the game  is described in Algorithm \ref{alg:agent}. {This agent implementation is the one that was delivered to participate in the Hearthstone AI competition. The values of the weights ${\mathbf w_1},\ldots, {\mathbf w_{21}}$ were calculated using the method described in next section.}
Note that the values given to {these} weights determine the importance of the distinct factors (e.g, hero health/armor, mana consumed, specific characteristics of the minions, etc) that {have any} influence on the value of $\Delta^{a,S{,\vec{w}}}$. Therefore, the assignment of values to these weights directly affects the {decision-making} mechanism of our agent. 
The source code of the agent is publicly available in our Github repository\footnote{\url{https://github.com/fergunet/SabberStone/tree/master/core-extensions/SabberStoneCoreAi/src/Agent}}.

\subsection{A Coevolutionary Approach to Optimize Hearthstone Agents}
\label{subsec:proposed-evolutionary-algorithm}

As described above, a Hearthstone agent is decision-making system driven by a collection $\langle \mathbf{w_1},\ldots,\mathbf{w_{21}} \rangle$ of 21 numerical coefficients, whose values ultimately dictate its behavior. Thus, increasing/decreasing the values of the weights may change the behavior of the agent, making it more aggressive, defensive, conservative, etc. It is in this context that the need {for} an optimizer becomes evident. The goal of this optimizer would be adjusting the weights so as to obtain a powerful, broadly successful strategy for playing Hearthstone. This optimization task will be tackled by means of EAs. More precisely, 
as the optimization process is conducted via the adjustment of  a real-coded vector with the weights of the score function for each action, the use of Evolution Strategies (ES) is proposed.  A general pseudo-code of the ES used in this work is reported in Algorithm \ref{alg:EA}.

\begin{algorithm}[!t]

\begin{algorithmic}
\STATE population $\gets$ initializeRandomPopulation() \COMMENT{{\em Create\ $\mu$\ individuals}}
\STATE evaluate(population)
\WHILE {stopping criterion not met}
    \STATE offspring $\gets$ mutate(population) \COMMENT{{\em Generate\ $\lambda$\ new\ individuals, including its associated $\sigma_i$}}
    \STATE evaluate(offspring + population)
    \STATE population $\gets$ population + offspring
    \STATE population $\gets$ reduce(population) \COMMENT{{\em Reduce population to initial size by removing worst individuals}}
\ENDWHILE
\end{algorithmic}

\caption{EA {for agent optimization}}
\label{alg:EA}

\end{algorithm}
As shown, the replacement mechanism chosen is a $(\mu + \lambda)$, that is, in every generation the best $\mu$ individuals out of the union set of the current $\mu$ parents and the newly-generated $\lambda$ offspring are kept as the parents for the next generation.

Individuals are represented as a vector $\vec{w} = \langle \mathbf{w_1},\ldots,\mathbf{w_{21}} \rangle$ of $n=21$ real values bounded to the $[0.0,1.0]$ range. Mutation is done using self-adapting non-correlated mutation amplitudes $\sigma_i$, $1\leqslant i \leqslant 21$. These mutation amplitudes are evolved along the genes of the candidates as described in Equations (\ref{eq:es1})-(\ref{eq:es3}):

\begin{align}
\sigma_i^\prime &= \max(\sigma_i + e^{\tau \cdot N_i(0, 1) + \tau^\prime \cdot N(0, 1)}, \epsilon)\label{eq:es1}
\\
\tau^{\phantom \prime} &= 1 / \sqrt{2 * \sqrt{n}}\label{eq:es2}
\\
\tau^\prime &= 1 / \sqrt{2 * n}\label{eq:es3}
\end{align}

Therein, $\sigma^\prime_i$ refers to the mutated value of $\sigma_i$, whereas $\tau$ and $\tau^{\prime}$ are the local and global learning rates, used as hyperparameters to control the self-adaptation of mutation amplitudes. In Equation (\ref{eq:es1}), the lower bound for any mutation amplitude $\sigma^\prime_i$ has been set to $\epsilon = 10^{-5}$. Once the mutation parameters have been updated, each variable 
$\mathbf{w_i}$ corresponding to each of the 21 weights of the agent is mutated as:

\begin{equation}
\mathbf{w_i} = \mathbf{w_i} + N(0, \sigma^\prime_i)
\end{equation}

Regarding the search space, it is important to stress that it has infinite size, as we are searching for the best combination of 21 real values that optimize the performance of our data-driven virtual player. 

When evolving a bot that plays a specific game, a fitness function needs to be properly defined to assess the efficiency of candidate solutions. A typical strategy to optimize game AI consists of matching each candidate solution against other efficient game AIs. Using an EA to execute this type of optimization is a natural way to obtain (with a high probability) a good solution (or at least a solution that can be considered acceptable). 

However, the problem becomes harder if no other AI for the game exists, or if no AI is available. For instance, simply because there is no proposal reported in the scientific literature, or because it is hard or even impossible to implement it due to the lack of specific details, just to name a couple of cases. Under these circumstances, the objective to optimize becomes unclear. Not having bots to ``spar'' with or improve against, may lead to underperformance against a wide variety of opponents, {such as} the ones expected in competitive tournaments. 

A possible way to deal with this issue is to use a Coevolutionary Algorithm (CoEA). As already mentioned in Section \ref{subsect: EA and CEA}, CoEAs are based on a {\em relative} approach to fitness evaluation. 
While standard EAs (already extensively used in the field of videogames, mainly for the automatic generation and refinement of AI engines \cite{MoraFGGF12,Spronck_OfflineEAs_03,Cole04GAFPS,Agent_Smith_CEC2009,GarciaSanchez2015}) are based on a predefined fitness function  (e.g., comprising a fixed set of opponents) that makes it possible to measure the quality of the candidate,  our CoEA is based on a {\em self-sustained} approach to fitness evaluation. That is, individuals are evaluated based on their interactions with other individuals. While in {\em cooperative} CoEAs the individuals work together to solve a problem (for example, each individual focusing on a specific part of the problem), in {\em competitive} CoEAs individuals are rewarded at the expense of the confrontation with other individuals. For example, Nogueira Collazo et al. \cite{nogueira14virtual} used a coevolutionary approach enriched with a hall-of-fame mechanism in order to retain a persistent memory of good strategies for a Real-Time Strategy game. They also took this approach one step further to coevolve both player agents and maps in a competitive fashion \cite{nogueira16competitive}.

In the case study described in this paper, the fitness of an individual is the number of victories it is able to obtain against the other members of the population and the freshly generated offspring. That is, for each of the $\mu$ individuals in the population and each of the new $\lambda$ candidates generated in the offspring, a number of games $g$ is performed for each combination of available decks {($D$)} and used to calculate the number of victories of the $\mu+\lambda$ members.
{To be precise, let $v\left( i,j,d_i,d_j,g\right)$ be a function that returns the number of victories (within range $[0,g]$) of the agent $i$ using the deck $d_i$ against the agent $j$ using the deck $d_j$, after playing $g$ games. Then, the fitness of an individual $i$ can be defined as:
\begin{equation}
    \mathit{Fitness}_{i}=\sum _{\substack{j\in \left( \mu +\lambda\right)\\ i\neq j} }\sum _{d_i,d_j\in D }v\left( i,j,d_i,d_j,g\right).
\end{equation}
Note that each agent in the population is represented as a collection of 21 real-valued weights that parameterizes the behavior of the decision-making procedure depicted in Algorithm \ref{alg:agent}, and where the values given to the weights have strong influence the decision of the best action to take. The performance of the agent in the game heavily depends on the sequence of the best actions taken by the agent during the gameplay in each turn. This surely affects the final result of the game (i.e., victory or defeat), which is then used within the fitness function as shown above.
}

The objective of this approach is to obtain more general bots than those evolved using a fixed set of opponents. In fact, when this particular agent for Hearthstone was evolved, no other agents were available online, so that is also a plus. Moreover, re-evaluating the parents at each generation reduces noise, and keeping the best individuals by having them compete against new candidate solutions that were not used before, makes it very difficult that suboptimal individuals survive. 

\section{Experimental evaluation}
\label{sec:experimental}
This section describes the experimental setup, the parameters used for the proposed approach, as well as the fitness evaluation used. First, the Hearthstone simulation engine used for this specific case study, called SabberStone, is described; then, the decks used for the fitness evaluation are summarized, and the proposed fitness function is presented. The {parameter} set for the EA and the hardware setup are described at the end of this section.

% -------------------------------------------

\subsection{Hearthstone Environment and Settings}
\label{subsec:metastone}
SabberStone \cite{SABBERSTONE} is an open-source (AGPLV3 licensed) Hearthstone simulator developed in C\#. It makes it possible to create agents and simulate all the aspects of Hearthstone, and it can be executed via command line. It is the chosen simulator for the CIG2018 Hearthstone AI competition \cite{competition18}.

Due to the rules of the CIG2018 competition, three pre-made decks are alternatively used by agents: an \emph{Aggro Pirate Warrior}, a \emph{MidRange Jade Shaman}, and a \emph{RenoKazakus Mage}. All decks are human-created by competitive players, and were prominently featured during previous seasons. In the following, we present a short expert analysis for each one, provided by one of the authors, that played Hearthstone since the Beta, and has currently over 14,000 recorded victories. The complete decklists are reported in Table~\ref{table:decklists}.

\subsubsection{Aggro Pirate Warrior (APW)}
Among the three decks, \emph{Pirate Warrior} seems by far the easiest to play for a human: when it was popular in ladder, this typology of aggressive decks attracted widespread criticism for requiring little skill to obtain good winrates\footnote{\emph{Pirate Warrior is Retarded}, discussion on Blizzard's official forums, \url{https://eu.battle.net/forums/en/Hearthstone/topic/17614485315}}. The deck has been structured to directly attack the opponent, reducing their hitpoints to zero as quickly as possible. For this reason, \emph{Pirate Warrior} features minions with Charge, able to attack immediately, like \emph{Patches the Pirate}, \emph{Kor'kron Elite}, \emph{Southsea Deckhand}; weapons like \emph{Fiery War Axe}, \emph{Arcanite Reaper}; and cards that power up or have considerable synergy with weapons, like \emph{Upgrade!}, \emph{Naga Corsair}, \emph{Bloodsail Raider}, \emph{Dread Corsair}. The most interesting choices for a player using this deck are selecting a new Hero power when \emph{Sir Finley Mrrgglton} enters play, and, maybe most importantly, deciding on how to deal with enemy Taunt minions: destroying them using weapon charges or trading with creatures might have a substantial impact on the final outcome of a game.

\subsubsection{MidRange Jade Shaman (MJS)} 
Another aggressive deck, this Shaman falls into the \emph{MidRange} category, as it exploits more powerful but more expensive cards than other aggro decks (see \emph{Pirate Warrior}), and makes ample use of the \emph{Jade} cards (\emph{Aya Blackpaw}, \emph{Jade Claws}, \emph{Jade Lightning}), that create minions of increasing strength when played. Unlike \emph{Pirate Warrior}, this deck features removal in the form of direct damage with cards like \emph{Maelstrom Portal}, \emph{Lightning Storm}, \emph{Lightning Bolt}, \emph{Jade Lightning}, and creatures with an excellent ratio between casting cost and statistics, like \emph{Tunnel Trogg}, \emph{Totem Golem}, \emph{Thing from Below}, \emph{Azure Drake}. A human player using this deck will face interesting choices, as several spells can often be used both as removal and as direct damage to the opponent. Moreover, there are several synergies between cards such as \emph{Azure Drake}, \emph{Bloodmage Thalnos}, all damaging spells, and \emph{Spirit Claws}, so a player might be torn between playing a card immediately to get an advantage, or wait and hope to draw a more powerful combination. Every choice is made more complex by the \emph{Overload} game mechanic, unique to Shamans, that makes it possible to play powerful cards on the current turn, but suffer a penalty in the next turn, in the form of a few temporarily unusable mana crystals.

\subsubsection{RenoKazakus Mage (RKM)}
Differently from the previous two decks, this \emph{Mage} list is built for control, aiming at removing early threats to then play extremely effective and costly cards in {mid-game}, to eventually win in the late game. Playing control is commonly believed to be more thought-intensive than playing aggression, and this deck is surely the most complex to play for a human, but there is another reason for that: the deck is built with only one copy of each card (while the limit is two copies), {thus making} every card both more unlikely to be available at the right time, and more precious, as it can be used only once. The decklist is structured around two cards whose \emph{Battelcry} effect triggers only if the player's deck contains at most one card per type: \emph{Reno Jackson}, that heals the player completely, and \emph{Kazakus}, that creates extremely powerful spells in the player's hand. The deck's strategy is to survive the early-to-mid game, using defensive cards such as \emph{Ice Barrier}, \emph{Ice Block}, \emph{Refreshment Vendor}, \emph{Mind Control Tech} to avoid death, and removals like \emph{Blizzard}, \emph{Volcanic Potion}, \emph{Flamestrike} to destroy large numbers of the opponent's minions. The deck presents a large number of interesting choices, ranging from whether to play removals or wait to obtain a better cost-effectiveness, with the risk of receiving more damage; to playing powerful cards immediately, or waiting for synergy (for example \emph{Bronn Bronzebeard} duplicates the effect of the player's battlecries, extremely important for \emph{Kazakus} and other minions such as \emph{Kabal Courier} or \emph{Azure Drake}); to evaluating the correct moment to heal completely using \emph{Reno Jackson}. As the rules of the competition do not allow the agents to mulligan their first hand, this deck is likely to be the hardest to play and the one with the lowest winrate.

\subsection{Experimental Setting}
The \texttt{inspyred} framework \cite{inspyred}, implemented in Python, has been used to develop the coevolutionary algorithm described in Section \ref{subsec:proposed-evolutionary-algorithm}. Source code of the algorithm is also available in our Github repository\footnote{\url{https://github.com/fergunet/SabberStone/blob/master/core-extensions/SabberStoneCoreAi/coevolutionary.py}}.

The coevolutionary algorithm has been configured with the parameters reported in Table \ref{table:ugp3} for all the experiments. {The parameters chosen for the ES are those used in similar works (e.g., they are the same as those used in the method described in \cite{Garcia18Knowsys})}. As the algorithm features stochastic elements, it has been executed 10 times ($E$) to perform a more reliable analysis. 

\begin{table}[htb]
\caption{Parameters used by the ES.}
\centering
\resizebox{12cm}{!}{
\begin{tabular}{ccc}
\hline
\textbf{Parameter} & \textbf{Meaning} & \textbf{Value}\\
\hline
$\mu$ & Population size & 10\\
$\lambda$ & Offspring size & 10\\
$\mathcal{G}$ & Number of generations &  100\\
$Str$ & Strategy & $(\mu+\lambda)$\\
$R$ & Replacement mechanism & Elitism \\
$e$ & Number of tested decks in each evaluation & 3 \\
$t$  & Number of games (per deck combination) in each evaluation & 20 \\
$E$ & Number of executions of the algorithm & 10 \\
\hline
\end{tabular}
}
\label{table:ugp3}
\end{table}

All the experiments have been executed on a computer with Intel Core i7-4770 CPU @ 3.40GHz $\times$ 8 processor, 32 GB RAM and Ubuntu 16.04. Due to the large number of games, deck combinations and individuals, each complete evolution required approximately a couple of days.

\section{Results and Discussion} 
\label{sec:results}

In this section we discuss different aspects of the proposed approach, such as the evolution of the individuals, the distribution of the obtained weights, the performance of the use of different decks, and the results of the best agent we obtained in the CIG2018 Hearhtstone AI Competition.

\subsection{Analysis of the Evolution}

As previously explained, the fitness of an individual depends on the number of victories against the other members of the population, {including its} parents. It is thus expected that the average fitness will tend to the half of the total number of games played by each individual of the offspring. 

\begin{figure}[!t]
\centering
\includegraphics[width=13cm]{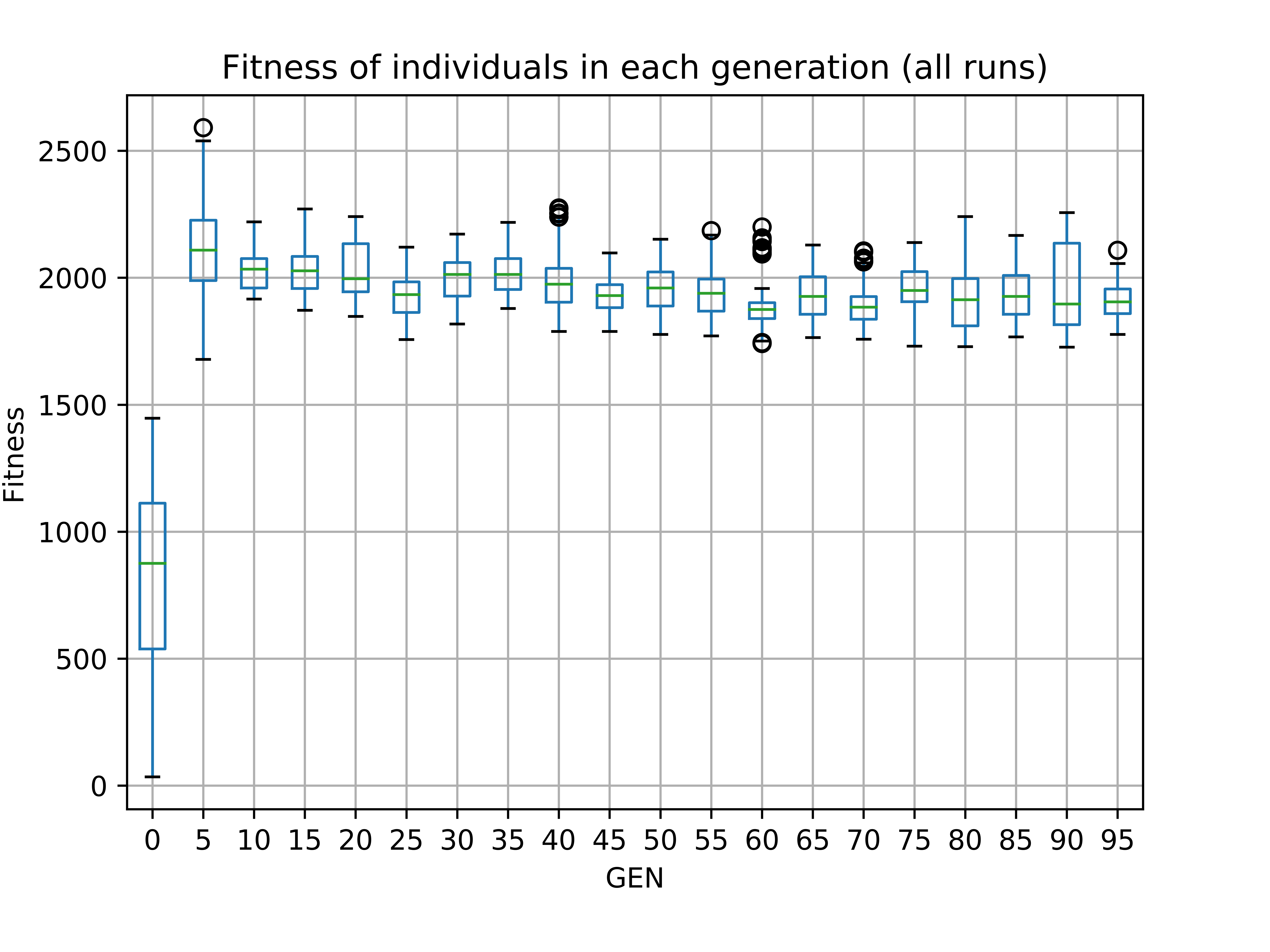}
\caption{Boxplots of the fitness distribution of the individuals in each generation (from all 10 ($E$) runs).}
\label{fig:fitness-gen}
\end{figure}

Figure \ref{fig:fitness-gen} shows the distribution of the fitness during the evaluation of all runs ($E=10$). 3420 victories is the maximum fitness for an individual that always wins  against all other individuals: 19 against individuals already in the population and the freshly produced offspring (not including the individual itself), 20 games, and 3$\times$3 decks combination = 3420. In generation 0 only the first randomly-generated individuals are evaluated against each other, therefore, the maximum fitness is limited to 1620. Random individuals can achieve 0 victories against the other individuals, and this generation shows a wide difference in fitness values. Generation 5 still features significant differences, but from Generation 10 onward, the differences are reduced, and the values seem stable until the end of the runs. {Although fitness might seem to have converged very quickly, it is necessary to point out that fitness values do not carry an absolute meaning as mentioned before, but reflect a relative figure of merit (because in each generation all the individuals --parents and offspring-- are again confronted with each other). In this sense and as an example, to obtain 2000 victories in the 50th generation is more difficult than in the 2nd generation, because new individuals aspiring to enter the population have to face tougher individuals than they did in that very early generation. Thus, the general population is improving over time, which is actually the purpose of the coevolutionary scheme.}

\begin{figure}[!t]
\centering
\includegraphics[width=13cm]{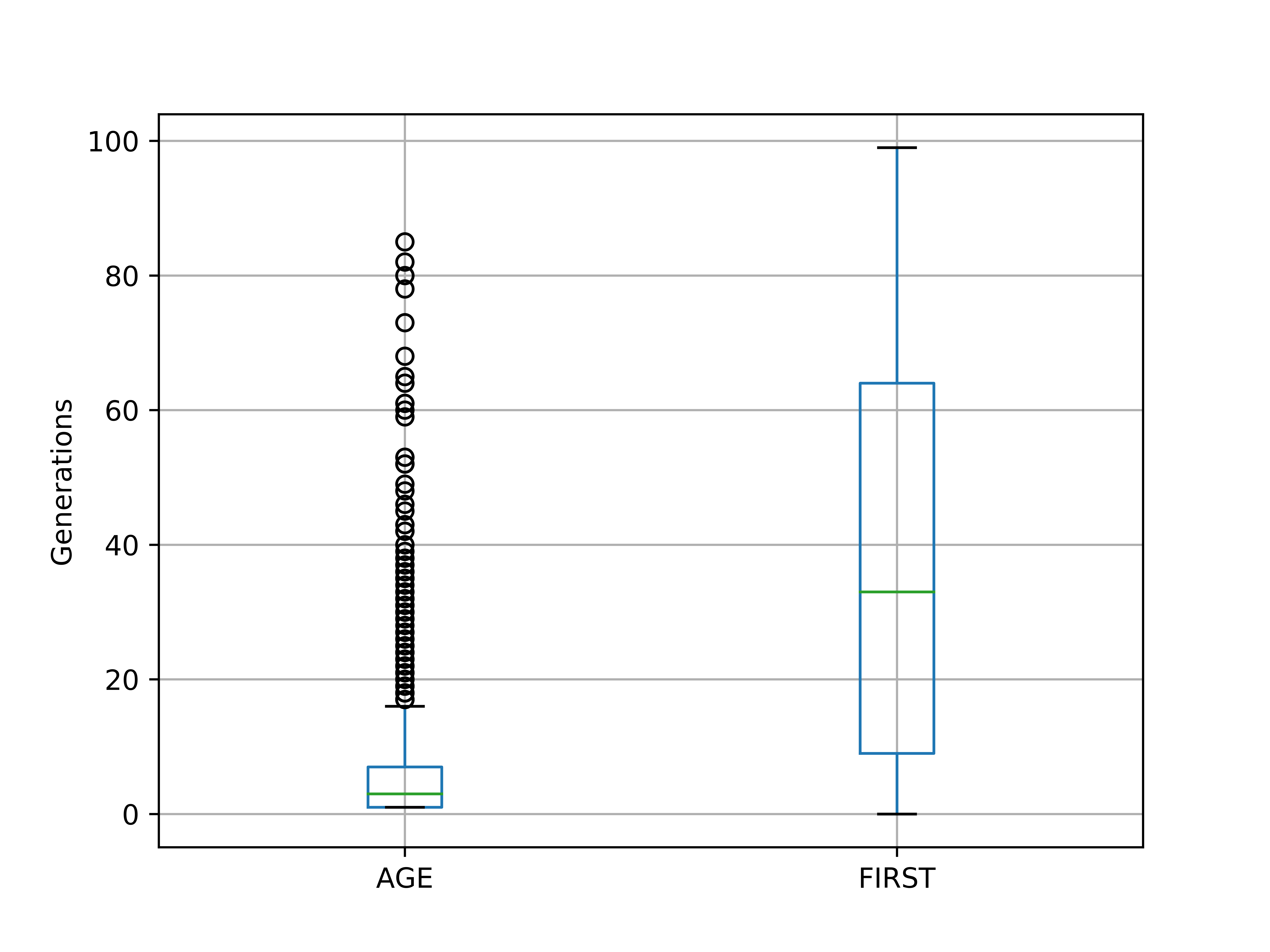}
\caption{Boxplots describing individuals' lifespan over all the 10 ($E$) runs. (AGE) describes the age of all individuals produced and (FIRST) 
shows the first generation that an individual entered the main population, thus being able to defeat at the very least the weakest individual in the previous population}
\label{fig:age}
\end{figure}

\begin{figure}[!t]
\centering
\includegraphics[width=13cm]{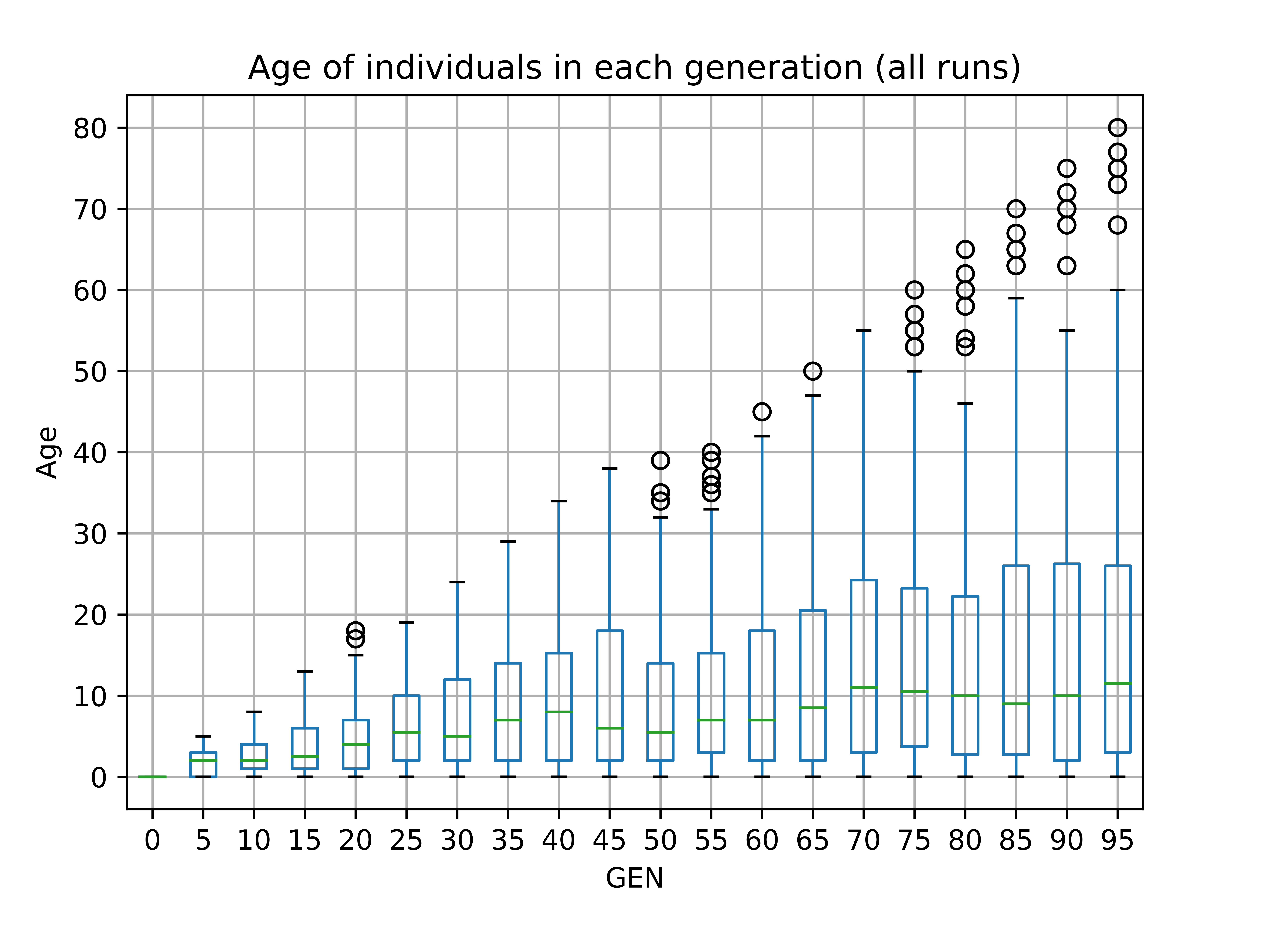}
\caption{Boxplots of the age distribution of all individuals in each generation (from all 10 ($E$) runs).}
\label{fig:age-gen}
\end{figure}

\begin{figure}[!t]
\centering
\includegraphics[width=13cm]{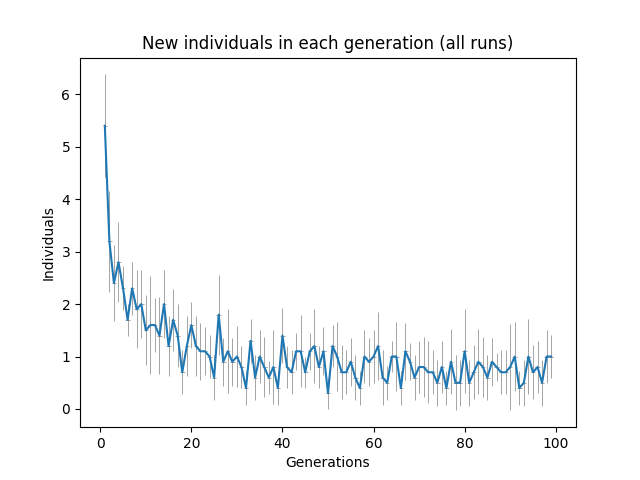}
\caption{Average number of new individuals that enter in the population in each generation (from all 10 ($E$) runs). {The lightgrey lines show 95\% confidence intervals.}}
\label{fig:newindividuals-gen}
\end{figure}

{As mentioned above, individuals are not evaluated only once: in each generation all parents and offspring are compared with each other again, so that the fitness of one parent may change in the next generation}. As previously stated, this not only avoids {the} presence of sub-optimal individuals that survived by chance, but also enforces keeping the best ones as long as they are strong enough. Figure \ref{fig:age} summarizes the age distribution of all individuals generated; and the distribution of the generation where an individual strong enough to survive appeared.  Half of the surviving individuals appeared between generations 30 and 60, which mark this part of the evolutionary process as the one featuring the most rapid improvement. The average age of individuals is around 5 generations, with several outliers distributed from 17 to 80 generations, meaning that strong individuals can also appear early in the evolution. This can be also seen in Figure \ref{fig:age-gen}, where these outliers appear after 50 generations. {The} average age of individuals is also increasing during the evolution, showing that fitter and fitter individuals are produced as the process goes on. However, this also implies that the number of new individuals that enter in the population decreases over time, as tougher individuals are more difficult to beat. Figure \ref{fig:newindividuals-gen} shows how the average number of new individuals that enter in the population, changes over generations, reaching a {stable} limit {value of} about 1 individual {on} average around generation 20.

\subsection{Analysis of the Solutions}
\begin{figure}[!t]
\centering
\includegraphics[width=13cm]{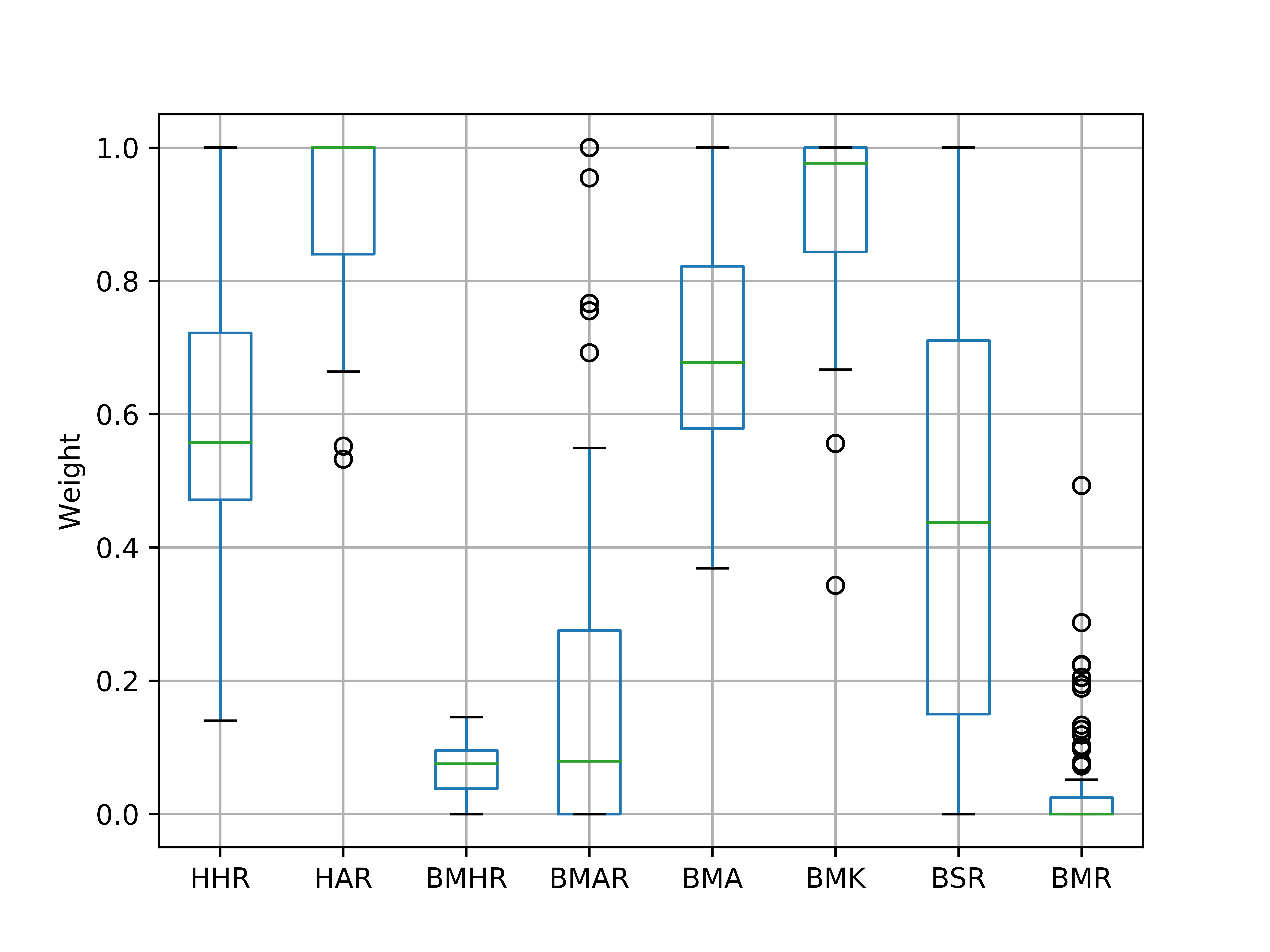}
\caption{Boxplots of the weight distribution that score the changes {on} the battlefield (the ones shown in Table \ref{tab:weightsBattlefield} used to calculate  
$\Delta^{a,S}_{attributes(hero)}$, $\Delta^{a,S}_{minions(hero)}$, $\Delta^{a,S}_{secrets(hero)}$ and $\Delta^{a,S}_{manaConsumed}$. 
These weight distributions have been obtained from all the individuals of the last generation of all runs.}
\label{fig:weights-battlefield}
\end{figure}

Analyzing the weights obtained from all 100 individuals inside the final population of 10 individuals at the end of each of 10 runs ($\mu \cdot E=100$) can provide further insight into the agents' behaviors. From Figure \ref{fig:weights-battlefield}, where the weights related {to} the changes of the battlefield are plotted, it can be seen that some weights have more variance than others.

The weight with the least variability in values is $BMR$ {(}i.e., $\mathbf{w_8}$ in Table \ref{tab:weightsBattlefield} and in Equation (\ref{eq:secretmana}){)}, being the one closer to 0. This might be  because in Equation (\ref{eq:score}) the score 
$\Delta^{a,S}_{manaConsumed}$
is negative. Therefore, actions that require less mana will score higher. This makes sense, because we are using a 
greedy policy that always {selects} the best action instead {of} simulating the best combination of actions (as in MCTS). It follows that from the equations it is preferable to perform a turn spending all mana performing several actions, than executing just one with higher mana cost but leaving some available mana unspent. For example, if the agent has 6 crystals available to spend in a specific turn, and a hand with three 2-cost cards and one 5-cost card, it is usually better to spend all the mana to perform 3 actions of cost 2, than one of cost 5, leaving 1 crystal without use. 

Another weight with smaller variation of values is $BMHR$ (minion health reduced, i.e., $\mathbf{w_3}$ in Table \ref{tab:weightsBattlefield}). It also makes sense that this weight tends to be low, while $BMK$ (minion killed{,} i.e., $\mathbf{w_6}$) tends to be high, so the selected actions are more oriented towards killing minions than injuring them. Injured minions are still almost as effective as healthy ones, so this weight reflects the common-sense notion that eradicating a minion from the board is strictly better than just damaging it. Similar weights are obtained by $HAR$ (Hero Attack {Reduced, i.e.,} $\mathbf{w_2}$), associated to card and actions that remove weapons from the enemy hero.

There are also weights with more variability than the previous ones. Weights that should have more importance, such as $HHR$ (changes in Hero Health i.e., $\mathbf{w_1}$) or $BSR$ (modification in secrets i.e., $\mathbf{w_7}$) show a higher degree of variability. It makes sense that $BSR$ (creation/destruction of a secret) does not importantly impact decisions, as only one deck of the three used in the training lists secrets, and only two cards ({\em Ice Barrier} and  {\em Ice Block}). Surprisingly, killing the enemy hero by reducing his health to 0 is the objective of the game, so it is curious that $HHR$ is not having average values closer to 1. This can be explained because agents are more focused to destroy minions, and the action to attack the enemy hero always will be executed when the desk is clear.

\begin{figure}[!t]
\centering
\includegraphics[width=13cm]{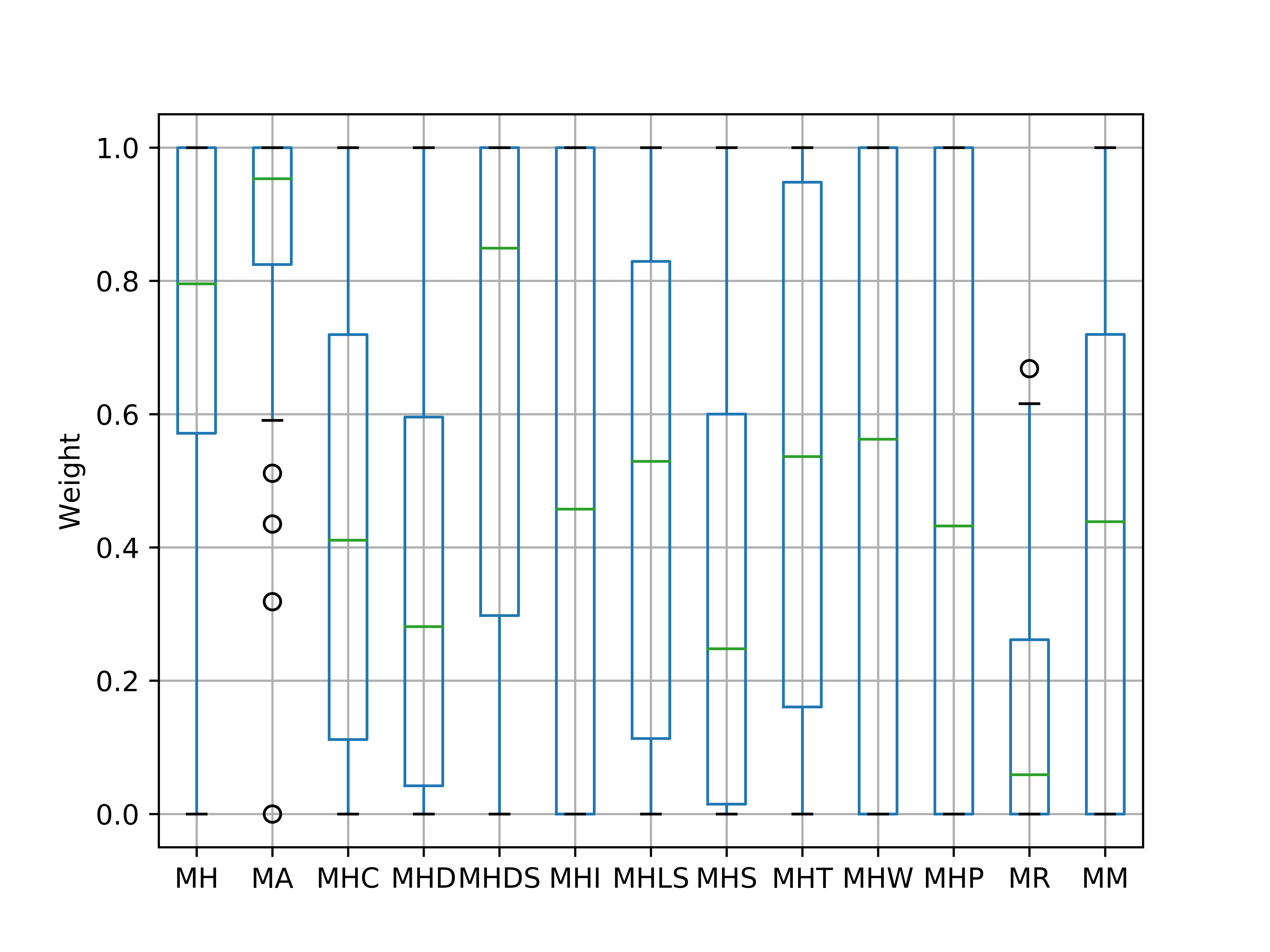}
\caption{Boxplots of the distributions of those weights that score a minion via the function $value\mathit{Of}(m)$ and that are shown in Table \ref{tab:weightsMinion}.
These distributions have been obtained from the 100 individuals at the end of all runs ($\mu\cdot E=10 \cdot 10$).}
\label{fig:weights-minions}
\end{figure}

As previously explained, scores also take into account the quality of the modified minion to be calculated by using the
function $value\mathit{Of}(m)$. The weights obtained from all individuals at {the} end of the run are plotted in Figure \ref{fig:weights-minions}. It is clear than some weights do not have so much influence, as their values are {commonly} distributed in all the range [0,1]: $MHI$, $MHW$, $MHP$ (namely, minions with Inspire, Windfury and Poison respectively i.e., $\mathbf{w_{14}}$, $\mathbf{w_{18}}$ and $\mathbf{w_{19}}$, respectively, in Table \ref{tab:weightsMinion}). Other weights also related {to} minion abilities are more distributed to lower values, such as $MHC$ (Charge), $MHD$ (Deathrattle), $MHS$ (Stealth) i.e., $\mathbf{w_{14}}$, $\mathbf{w_{11}}$ and $\mathbf{w_{16}}$, respectively, in Table \ref{tab:weightsMinion}. It is clear that the most important weights are the ones related {to} the Health ($MH$) and Attack ($MA$) i.e., $\mathbf{w_{9}}$ and $\mathbf{w_{10}}$ respectively. In fact, an action will be more rewarded if it summons (or attacks) minions with higher attack values. Moreover, even {though} {\em Rarity} is a commonly used way to distinguish ``good'' cards from ``weak'' ones, the $MR$ weight (i.e., $\mathbf{w_{20}}$) is the one closer to 0, implying that this statistic cannot be used as a proxy of a card's effectiveness in a specific play situation: knowing that a card is Rare, in other words, is not informative of whether it should be played on a particular board configuration.

\subsection{Deck behavior}

Table \ref{tab:percdecks} shows the percentage of wins of all individuals inside the final population at the end of all runs (100), separated by deck type. Note that these percentages also take into account the number of victories against the individuals in the population and offspring that did not pass the cut to survive the generation, so that is the reason why mirror matches report percentages other than 50\%. Overall, there is not much difference between matchups, although it is clear that the evolved individuals obtain the highest victory rates against the Mage deck, that is predictably the hardest to play for the AI agents, as previously discussed. 

\begin{table}[!t]
    \caption{Percentage of victories of the individuals at the and of the runs by deck. Row name wins against column name. Note that this victories also take into account the games against the other parents/offspring that did not survive the generation. }
    \centering
    \begin{tabular}{c|c|c|c}
	     & Warrior   &	Mage     &	Shaman \\ \hline
Warrior  &	57,72 \% & 	73,15 \% &	41,64 \% \\ \hline
Mage	 & 52,41  \% &	65,28 \% &	58,12 \% \\ \hline
Shaman   &	58,08 \% &	61,83 \% &	48,95 \% \\ \hline
    \end{tabular}
    \label{tab:percdecks}
\end{table}

\subsection{Characterization of Solutions}

\begin{figure}[!t]
\centering
\includegraphics[width=13cm]{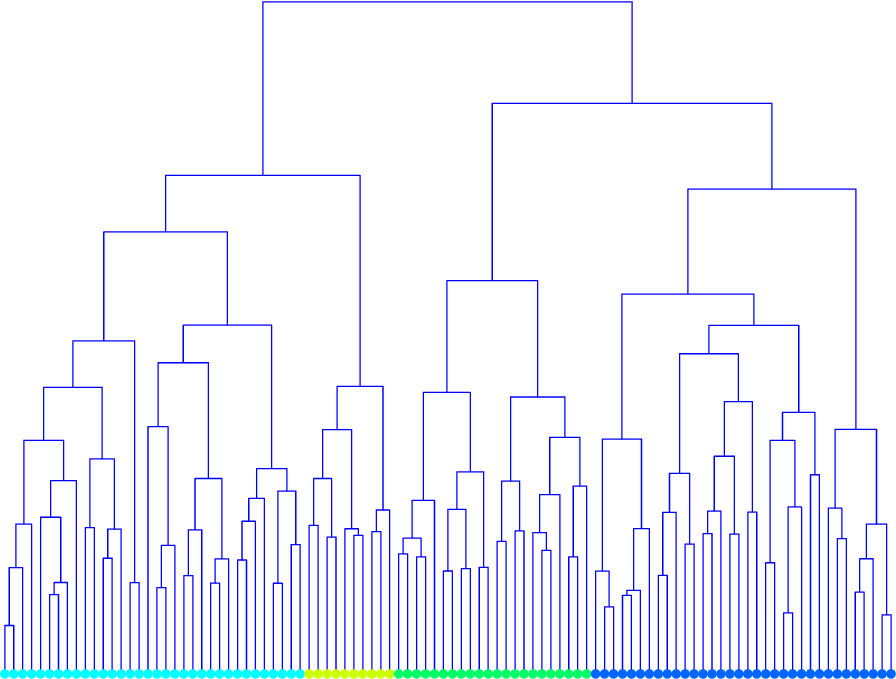}
\caption{Hierarchical clustering of solutions according to Ward algorithm. The color of the leaves corresponds to the clusters obtained following the Silhouette criterion.}
\label{fig:clustering}
\end{figure}

Let us now turn our attention to the typology of solutions provided by the algorithm. To this end, we have picked the last population in each run of the coevolutionary algorithm, thus creating a pool of the 100 solutions at the end of the runs. Each solution is described by a numerical vector of 21 values, hence indicating a point in a 21-dimensional space. We have considered the Euclidean distance between these points as a measure of dissimilarity among pairs of solutions. Solutions can then be grouped on the basis of this distance metric. To this end, we used the Ward algorithm for performing a hierarchical clustering of the solutions \cite{ward63hierarchical}. Subsequently, in order to determine a suitable partition of this tree into a number of groups, we have considered the Silhouette criterion \cite{Rousseeuw87silhouettes} at each level, resulting in four groups. The hierarchical clustering tree and the four groups identified are shown in Figure \ref{fig:clustering}. These groups are named \#1, \#2, \#3 and \#4 in the rest of the paper.

\begin{figure}[!ht]
\centering
\includegraphics[width=13cm]{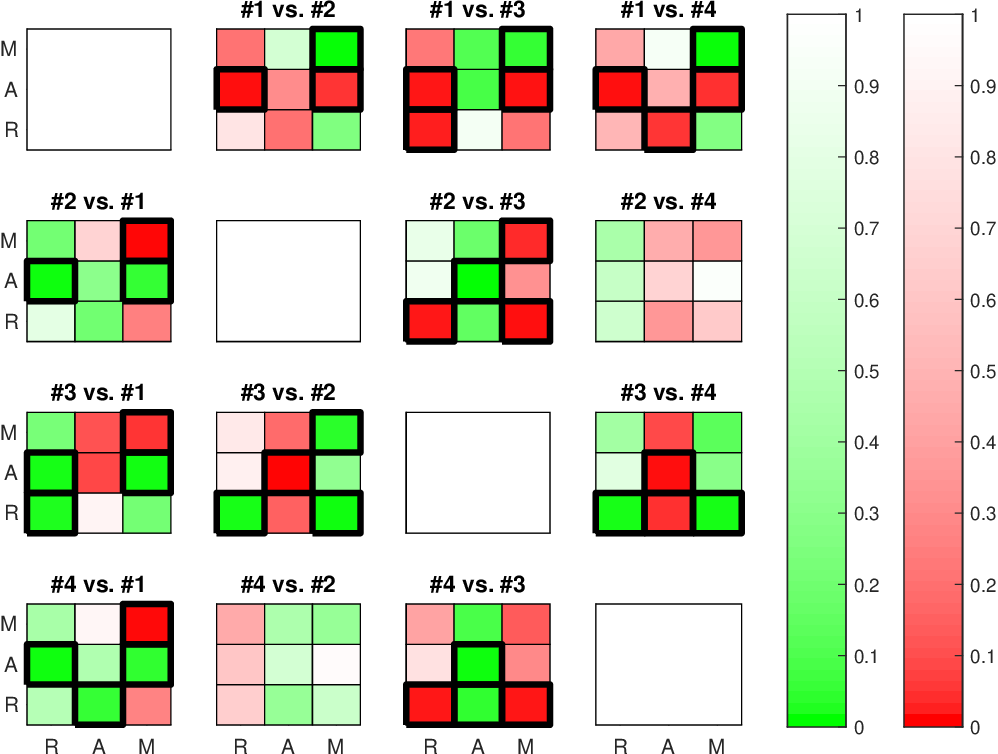}
\caption{Relative behavior of each group as a function of the decks used (\textbf{M}idrangeJadeShaman, \textbf{A}ggroPirateWarrior, \textbf{R}enoKazakusMage). Each subfigure corresponds to the differential behavior of solutions in two groups when confronted with an arbitrary opponent (the rows correspond to the decks used by the groups under scrutiny and the columns to the decks used by the opponent). Green (resp. red) shades indicate superiority of the first (resp. second) group. The intensity of the shade indicates the $p$-value of the head-to-head comparisons (see color bars). Statistically significant comparisons (at $\alpha=.05$) are shown by thick boxes.}
\label{fig:comparison1}
\end{figure}

\begin{figure}[!ht]
\centering
\includegraphics[width=13cm]{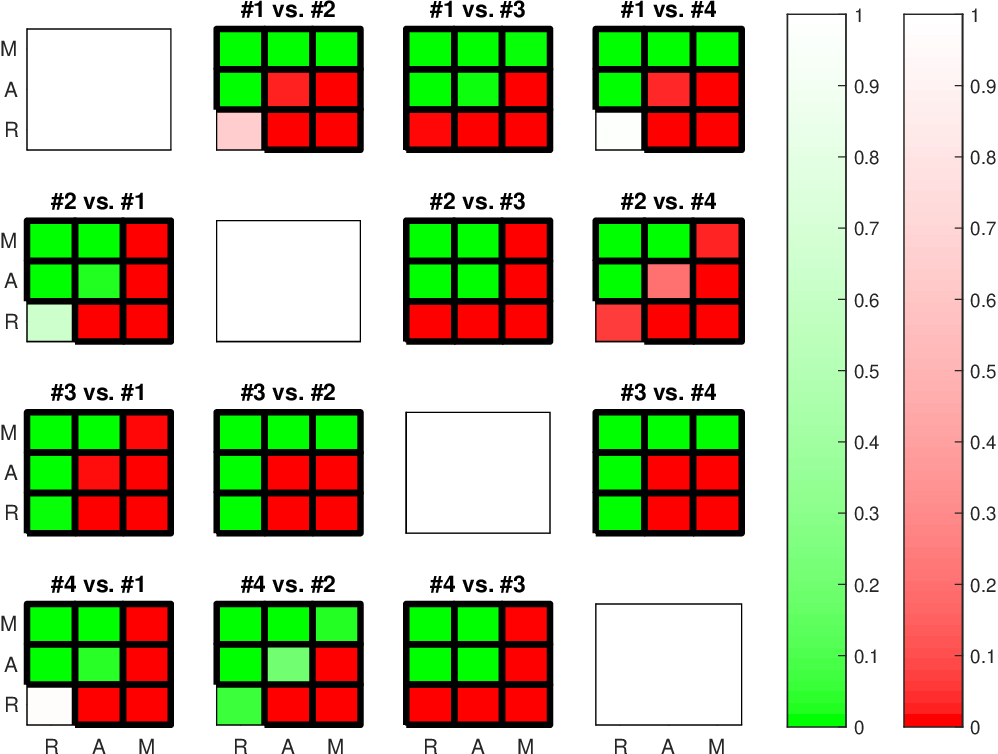}
\caption{Direct comparison of the behavior of each group as a function of the decks used. Each subfigure corresponds to the direct match-up between solutions in two groups (the rows correspond to the decks used by the first group and the columns to the decks used by the second group). Colors and boxes have the same meaning as in Figure \ref{fig:comparison1}.}
\label{fig:comparison2}
\end{figure}

\sloppypar We then consider the relative behavior of each group when playing with/against certain decks. For this purpose, we measure the number of battles won for each of the nine deck combinations (i.e., decks played by each player), comparing each solution in the pool with every other solution, for a total of ($100\cdot99/2)\cdot9 = 44,550$ match-ups. The distribution of these values across each group is tested against all other groups using a Wilcoxon ranksum test. This is performed in two complementary ways: (1) we compare the behavior of two groups $G$ and $G'$ when a solution from any of these groups plays with deck $d_1$ against any other solution in the pool (regardless of its group) with deck $d_2$; (2) we compare the behavior of two groups $G$ and $G'$ when a solution from $G$ plays with deck $d_1$ against a solution in $G'$ with deck $d_2$. Notice that the latter comparison is intended to analyze the direct head-to-head behavior of solutions in each group for each deck combination, whereas the former provides an illustration of the relative differential behavior among groups when they use the same deck against a certain opponent. The outcome is summarized in Figures \ref{fig:comparison1}-\ref{fig:comparison2}.

Figure \ref{fig:comparison1} presents the differences between combinations of deck/group when confronting an arbitrary opponent. This figure shows how individuals from group \#1 perform better than all the other groups only when using the Midrange Shaman against the same deck, but having the worse results when using the Pirate Warrior. On the contrary, groups \#2 and \#4 have the best results with Warrior, being \#4 the only that can manage to win using Pirate Warrior against the Mage deck. Group \#3 has a mixed performance, obtaining better results in combinations that are not Pirate Warrior vs Pirate Warrior. This can be corroborated by the results shown in Figure \ref{fig:comparison2}, where a direct match-up results can be checked: \#1 obtains the most of its fitness by using Shaman, \#2 and \#4 take the most advantage of the Pirate Warrior, while \#3 is the only group of agents that wins more times with Mage. However, as expected, the Mage deck can only win reliably against other Mage match-ups, as in the case of \#3.

\subsection{Hearthstone AI competition results}
In August 2018, the first agent obtained from our method, nicknamed {\em EVA} (EVolutionary Agent), participated to the Hearthstone AI Competition held at CIG2018, in the ``Premade Deck Playing'' track. 33 agents were submitted to the competition, and its top performers are shown in Table \ref{tab:competition}. More information about this event is available in a presentation on the competition's website \cite{competition18} and in \cite{Dockhorn2019}. Our evolved agent ended up in second position, out of the 33 presented. It must be taken into account that the solution sent to the competition was the best individual from the first completed evolutionary process, not the best one obtained from all runs, due to time constraints regarding the competition deadline. Interestingly, our approach even defeats agents that use MCTS, the state-of-the-art \cite{Swiechowski18} for tree search in card games. {We hypothesize the reason for this is manifold. On one hand, MCTS agents were not specifically optimized to use the competition decks, unlike our proposal. On the other hand, MCTS usually utilizes a user-defined heuristic to select a tree node and a high-quality set of configuration parameters, which may not be the most appropriate, as these were not evaluated against an adequate set of opponents with different behaviors, as in our case.}

\begin{table}[htb]
\caption{Winners of the ``Premade Deck Playing'' track of the CIG2018 Heartstone AI competition (out of 33 participants).}
\label{tab:competition}
\centering
\begin{tabular}{c|c|c}
   Rank  & Name (Method, if known) & Winrate \\ \hline
   1     & Max Frick, Unal Akkaya (?) & 76.0\% \\
   2     & \textbf{EVA (Optimized Greedy)} & \textbf{74.2} \% \\
   3     & Kai Bornemann (MCTS) & 72.5 \% \\
   4     & Hans-Martin Wulfmeyer (Alpha-Beta Pruning) & 68.3\% \\
   5     & Ivan Prymak, Milena Malysheva (?) & 68\% \\ \hline
\end{tabular}
\end{table}

\begin{figure}[htb]
\centering
\includegraphics[width=13cm]{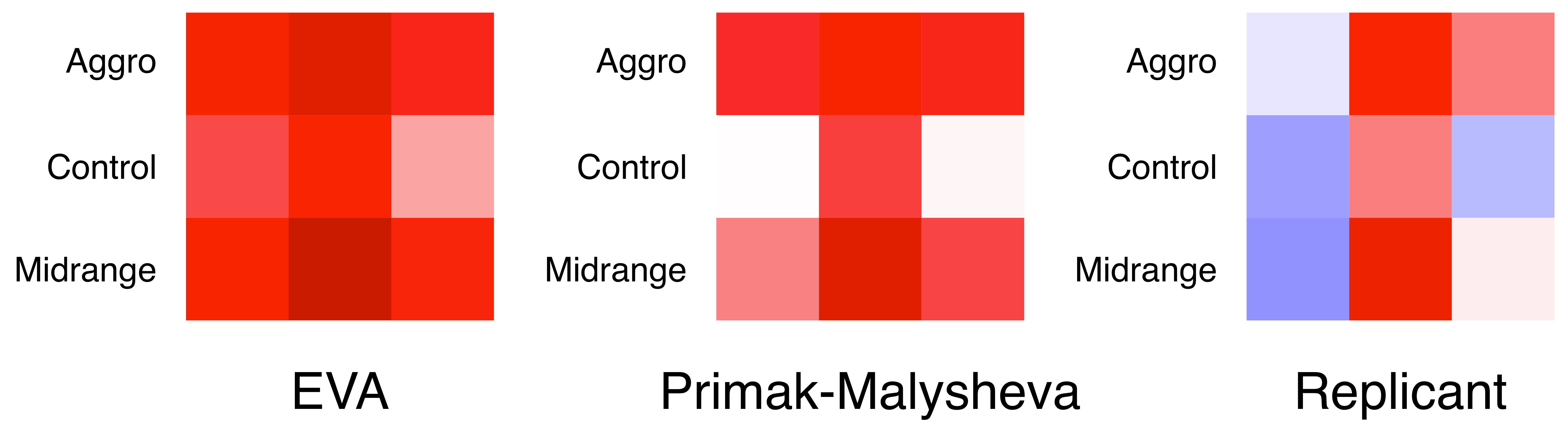}
\caption{Winrate of each deck configuration for three bots during the CIG2018 Competition: EVA, Prymak-Malysheva and Replicant. Winrates are shown in color intensity, from darker blue (0\% winrate) to darker red (100\% winrate), being white the 50\%. Colours obtained from \cite{competition18}}.
\label{fig:competitioncolors}
\end{figure}

From the aforementioned presentation given at the end of the competition \cite{competition18} we have extracted the decks winrate results shown in Figure \ref{fig:competitioncolors}, describing the behavior of our agent with respect to the one that ended with a lower score in 5th position (Prymak-Malysheva) and against an agent that did not end in the top 5 (Replicant). As it can be seen, our agent performs well for all possible deck combinations, obtaining more than 50\% of winrate in all of them, similar to the results shown in Figure \ref{tab:percdecks}. The other bots show some differences in winrates: Prymak-Malysheva has some difficulties mastering Mage, and Replicant only obtains good results (positive winrates) against Mage decks. With respect to the other winning agents, it is mentioned in \cite{competition18} that best bots performed well in all configurations.

\section{Conclusions}
\label{sec:conclusions}
In this work we demonstrate that using competitive coevolutionary optimization allows to obtain agents that play the Hearthstone game with enough versatility to play using/against different decks. An evolutionary strategy (ES) has been used to optimize  the value of 21 parameters that lead the decision-making mechanism to select the best action to play at each moment of the player turn. 
Moreover, the proposed method does not require any existing ``parry'' bot for fitness evaluation to improve against, but the rest of the population is used to calculate its performance during the evolution. Furthermore, one of the generated agents by our method finished second in the Hearthstone AI Competition 2018 \cite{competition18}, obtaining good winrates in every deck configuration.

Our evolved agents show differences in performance due to the variation of the values associated to the weights that guide the game artificial intelligence. Thus, a clustering of the solutions has been performed to characterize the generated agents, {analyzing} the relative {behavior} of the different groups obtained depending on the decks used. 
In addition, an analysis {of} the influence that each factor involved in the function that guides the selection of actions to take, has been addressed.

It must be noted that our proposal only takes into account one possible action to execute from the current state. Considering more possible actions (perhaps even in sequence), 
while requiring more computational time, could improve the performance of the agent, as the Greedy turn-based agents {outperform} the movement-based ones (and note that an `action' is  considered a `movement') \cite{Garcia18Knowsys}. In fact, some MCTS method can be included in our proposal in future studies.

Moreover, one of the drawbacks of our previous work \cite{Garcia18Knowsys} is that it was limited to the performance of the fixed AI available in MetaStone for the evolution of the deck. In future {work}, we can study the combination of collaborative-competitive coevolutionary methods to use different populations, each one aimed to a different objective: one to improve the parameters of the agent for a specific deck, as we did in this work, and other devoted to {generating} the decks to play (as done in \cite{Garcia18Knowsys}), in order to obtain agent/decks configurations that best suit the current state of the evolution. This might lead to {obtaining} good agents to participate in the ``{User-created} deck playing'' track in future competitions, to demonstrate the feasibility of the coevolutionary competitive-collaborative methods in this area.

\section*{Acknowledgements}
\sloppypar This work has been partially funded by projects SPIP2017-02116, EphemeCH (TIN2014-56494-C4-\{1,3\}-P), DeepBio (TIN2017-85727-C4-\{1,2\}-P) and TEC2015-68752 and ``Ayuda del Programa de Fomento e Impulso de la actividad Investigadora de la Universidad de C\'adiz''. 

\bibliographystyle{elsarticle-num}
\bibliography{hearthstone}

\appendix
\section{CIG2018 Hearthstone AI Competition Decklists}
Table \ref{table:decklists} show the decklists of the used decks for evaluation, that were proposed in the CIG2018 Hearthstone AI challenge.

\begin{table}[htb]
	\centering
	\caption{Decklists used in the competition. Cards are reported in the order they were presented in the deck's text files.}
	\label{table:decklists}
	\resizebox{\textwidth}{!}{%
		\begin{tabular}{l|l|l}
			\multicolumn{1}{c|}{\textbf{Pirate Warrior}} & \multicolumn{1}{|c|}{\textbf{Aggro Shaman}} & \multicolumn{1}{|c}{\textbf{Control Mage}} \\ \hline
			Sir Finley Mrrgglton                        & Tunnel Trogg                              & Forbidden Flame                           \\
			Fiery War Axe                               & Tunnel Trogg                              & Arcane Blast                              \\
			Fiery War Axe                               & Totem Golem                               & Babbling Book                             \\
			Heroic Strike                               & Totem Golem                               & Frostbolt                                 \\
			Heroic Strike                               & Thing from Below                          & Arcane Intellect                          \\
			N'Zoth's First Mate                         & Thing from Below                          & Forgotten Torch                           \\
			N'Zoth's First Mate                         & Spirit Claws                              & Ice Barrier                               \\
			Upgrade!                                    & Spirit Claws                              & Ice Block                                 \\
			Upgrade!                                    & Maelstrom Portal                          & Manic Soulcaster                          \\
			Bloodsail Cultist                           & Maelstrom Portal                          & Volcanic Potion                           \\
			Bloodsail Cultist                           & Lightning Storm                           & Fireball                                  \\
			Frothing Berserker                          & Lightning Bolt                            & Polymorph                                 \\
			Frothing Berserker                          & Jade Lightning                            & Water Elemental                           \\
			Kor'kron Elite                              & Jade Lightning                            & Cabalist's Tome                           \\
			Kor'kron Elite                              & Jade Claws                                & Blizzard                                  \\
			Arcanite Reaper                             & Jade Claws                                & Firelands Portal                          \\
			Arcanite Reaper                             & Hex                                       & Flamestrike                               \\
			Patches the Pirate                          & Hex                                       & Acidic Swamp Ooze                         \\
			Small-Time Buccaneer                        & Flametongue Totem                         & Bloodmage Thalnos                         \\
			Small-Time Buccaneer                        & Flametongue Totem                         & Dirty Rat                                 \\
			Southsea Deckhand                           & Al'Akir the Windlord                      & Doomsayer                                 \\
			Southsea Deckhand                           & Patches the Pirate                        & Brann Bronzebeard                         \\
			Bloodsail Raider                            & Small-Time Buccaneer                      & Kabal Courier                             \\
			Bloodsail Raider                            & Small-Time Buccaneer                      & Mind Control Tech                         \\
			Southsea Captain                            & Bloodmage Thalnos                         & Kazakus                                   \\
			Southsea Captain                            & Barnes                                    & Refreshment Vendor                        \\
			Dread Corsair                               & Azure Drake                               & Azure Drake                               \\
			Dread Corsair                               & Azure Drake                               & Reno Jackson                              \\
			Naga Corsair                                & Aya Blackpaw                              & Sylvanas Windrunner                       \\
			Naga Corsair                                & Ragnaros the Firelord                     & Alexstrasza                               \\
			&                                           &                                          
		\end{tabular}%
	}
\end{table}

\end{document}